\documentclass[journal]{IEEEtran}
\usepackage[T1]{fontenc}
\usepackage[utf8]{inputenc}
\usepackage{lmodern}
\usepackage{amsmath,amssymb,bm}
\usepackage{graphicx}
\usepackage{booktabs}
\usepackage{dblfloatfix}
\usepackage{float}
\usepackage[section]{placeins}
\usepackage{cite}
\usepackage{xcolor}
\usepackage[hidelinks, pdfauthor={Hyung Gon Shin, Sanghoon Jung}, pdfcreator={LaTeX}, pdfsubject={Research article}, pdftitle={Doing More with Less: Complex Manipulation with a Contact-Adaptive Robotic Hand}]{hyperref}

\newcommand{\newgreen}[1]{#1}
\begin{document}
\title{Doing More with Less: Complex Manipulation with a Contact-Adaptive Robotic Hand}
\author{Hyung Gon Shin$^{*}$ and Sanghoon Jung%
\thanks{*Corresponding author: Hyung Gon Shin. E-mail: \href{mailto:tlsgudrhs@postech.ac.kr}{tlsgudrhs@postech.ac.kr}, \href{mailto:ko.h.g.shin@gmail.com}{ko.h.g.shin@gmail.com}, and \href{mailto:hg0310.shin@samsung.com}{hg0310.shin@samsung.com}.}}
\maketitle
\begin{abstract}
Anthropomorphic robotic hands offer the potential to manipulate objects designed for human hands in environments adapted to human use. However, developing a robotic hand that replicates the human hand in terms of morphology, degrees of freedom, force output, durability, and sensing capabilities remains a formidable challenge, as these requirements inherently involve trade-offs. Here, we propose a robotic hand mechanism designed to address several of these trade-offs by enabling contact-adaptive finger motion, slender fingers, a wide grasping range, and sustained gripping force. We also introduce a motion glove based on quaternion swing--twist decomposition to extract human manipulation intentions and translate them into robotic hand commands, along with a forearm-mounted interface designed to enhance the operator's sense of embodiment. Experimental demonstrations show that a robotic hand with limited active degrees of freedom and no tactile sensing can perform complex tasks, including smartphone pickup and charging, in-hand manipulation, object unboxing and repackaging, and cutting and tying cords, through the combination of the proposed mechanism and effective bimanual coordination.
\end{abstract}
\section{Introduction}

Conventional robotic automation has often relied on task-specific grippers designed for particular objects and environments configured for robots \cite{ref1,ref2}. However, many tasks performed by humans involve a wide variety of objects in environments already designed for human use. In this context, anthropomorphic robotic hands offer the potential for versatile manipulation of everyday objects in human-centered environments with fewer task-specific modifications \cite{ref4,ref5,ref7,ref8}. Furthermore, their morphological similarity to human hands can facilitate motion retargeting when applying learning from demonstration (LfD) to robotic manipulation \cite{ref13,ref14,ref15}.

Although the design of anthropomorphic robotic hands involves many factors, here we focus on five key requirements: (i) a large number of degrees of freedom (DOF), (ii) high force output, (iii) high durability, (iv) rich sensing capabilities, and (v) shape and size comparable to those of the human hand. Satisfying all five requirements simultaneously remains a formidable challenge, as trade-offs among them are difficult to avoid in practice.

Representative trade-offs arise when increasing active DOF in a fixed volume requires smaller motors, reducing force output and heat dissipation and potentially limiting continuous operation \cite{ref16,ref17}. High-ratio transmissions can increase joint torque but reduce backdrivability and impact tolerance \cite{ref19,ref21}. Other considerations include structural strength, surface materials, sealing, sensing, and cost. Balancing these requirements motivates alternative hand mechanisms.

Several approaches have been proposed to address these trade-offs. Tendon-driven hands typically place actuators in the palm or forearm and transmit force to the finger joints through cables \cite{ref22,ref23,ref24,ref25}. This arrangement allows a large number of degrees of freedom within a compact space while accommodating relatively large motors with low gear ratios. However, tendon compliance, cable slack, and mechanical coupling can introduce transmission errors and complicate precise control \cite{ref28,ref29}.

Motor-integrated hands place actuators within or near the finger joints \cite{ref30,ref32}. This allows independent joint control and facilitates modular design and maintenance. However, the limited space within the joints restricts motor size and heat dissipation, potentially reducing force output and continuous operating time. High-ratio transmissions can also reduce backdrivability, making the joints more vulnerable to external impacts. Recent studies have explored methods to improve thermal management \cite{ref35,ref36}.

Linkage-driven hands transmit force from palm-mounted actuators through rigid links \cite{ref8,ref42,ref43}. Multiple linkages provide independent joint motion, whereas underactuated arrangements offer contact adaptation with fewer motors. They can provide high force transmission and mechanical robustness, but their transmission ratios may vary with joint configuration, and singularities may also occur \cite{ref42,ref45}. Link motion also occupies space and complicates enclosed finger covers.

Other approaches have also been proposed, including hydraulically actuated hands, soft robotic hands, and hands combining multiple mechanisms \cite{ref46,ref48,ref50}.

\newgreen{Another challenge in developing anthropomorphic robotic hands is how to clearly evaluate their manipulation performance. In most applications, robotic hands are mounted on articulated robot arms or integrated into humanoid robots rather than used independently \cite{ref52,ref13}. Consequently, manipulation outcomes depend on various factors beyond the hand itself, including robot arm performance and capabilities for bimanual coordination, perception, and decision-making. This makes it difficult to isolate the hand's contribution to manipulation performance. Since anthropomorphic hand design involves inherent trade-offs, clearly characterizing the hand's own capabilities and limitations can help identify an appropriate balance among competing design requirements.}

In this paper, we present four main contributions. \newgreen{First, we propose a contact-adaptive finger mechanism designed to address several of these trade-offs. Its unique split-stage gear--spring architecture enables a three-joint underactuated mechanism to be integrated into slender, fully enclosed, human-like fingers. With a single actuator per finger, the joints adapt mechanically to object contact, translating simple control inputs into contact-dependent finger configurations without independent joint actuation or tactile sensing. The reduced actuator count allows relatively large motors for sustained gripping force during continuous operation. The gear-based transmission allows a wide range of motion at each joint while maintaining constant transmission ratios across joint configurations. Together, these features support not only versatile grasping but also complex manipulation tasks.}

\newgreen{Second, we develop a simplified mechanical model of the proposed underactuated finger and derive design conditions intended to suppress the closing-ejection phenomenon. In underactuated fingers, closing-ejection refers to an unintended loss of contact during finger closure that may cause the object to slip or be ejected \cite{ref42,ref43}. The derived conditions relate gear transmission ratios and finger geometry to the suppression of unintended joint flexion, providing a design basis for reducing the risk of closing-ejection under the assumed contact conditions.}

\newgreen{Third, we introduce the branch-shadowing teleoperation interface to evaluate the proposed hand with reduced dependence on other robotic subsystems. Designed to support the operator's sense of embodiment, the interface uses motion gloves for finger control while delegating perception, decision-making, and bimanual manipulation to the human operator. This arrangement provides insight into the proposed hand's potential and limitations when paired with advanced perception and decision-making in future autonomous robotic systems.}

Fourth, we propose a motion glove to generate control inputs that allow the proposed hand to follow the operator's intended finger movements. Because the proposed hand uses a minimal number of actuators, successful motion retargeting requires extracting only the finger rotation angles needed for control from the human hand's multi-DOF motion. The proposed motion glove uses six attitude and heading reference system (AHRS) sensors and quaternion swing--twist decomposition to extract these angles.

\begin{figure}[!t]
\centering
\includegraphics[width=\columnwidth,height=.38\textheight,keepaspectratio]{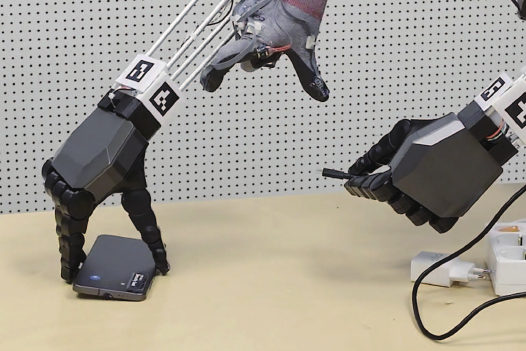}
\caption{Smartphone pickup and charging-cable insertion using the proposed robotic hands. The robotic hands use a contact-adaptive split-stage gear--spring mechanism and are controlled through motion gloves that generate finger control inputs from human hand movements.}
\label{fig:1}
\end{figure}

Using the proposed approaches, we demonstrate through quantitative and qualitative experiments that a robotic hand equipped with the proposed mechanism can perform complex and diverse manipulation tasks when combined with human-level bimanual coordination through the proposed teleoperation interface (Fig.~\ref{fig:1}).

\FloatBarrier
\section{Methods}

\subsection{Split-Stage Gear--Spring Mechanism}

\begin{figure*}[!t]
\centering
\includegraphics[width=\textwidth,height=.62\textheight,keepaspectratio]{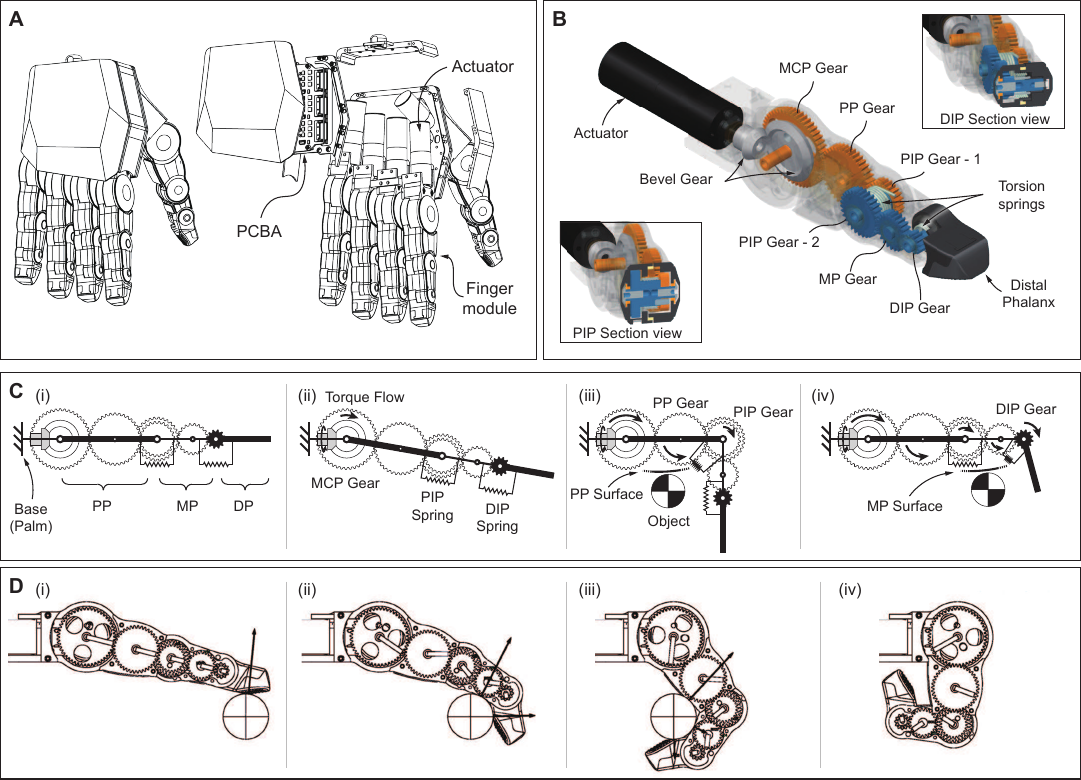}
\caption{Structure and contact-adaptive mechanism of the proposed robotic hand. \textbf{(A)} Overall structure of the robotic hand, consisting of five finger modules, actuators, and a printed circuit board assembly (PCBA). \textbf{(B)} Structure of the finger module. The split-stage gear--spring mechanism consists of two parallel spur gear stages and torsion springs at the proximal interphalangeal (PIP) and distal interphalangeal (DIP) joints. The gears are either fixed to the finger frames or mounted to rotate freely relative to them, enabling three-joint underactuated motion with a single actuator. \textbf{(C)} Mechanism of contact-adaptive finger behavior. The actuator torque is transmitted through the split-stage gear train, allowing the joints to adapt to different contact conditions without additional actuators or tactile sensing. (i) Simplified representation of the finger mechanism, including gears and torsion springs. (ii) Metacarpophalangeal (MCP) flexion before object contact. (iii) PIP flexion following contact with the proximal phalanx. (iv) DIP flexion following contact with the middle phalanx. \textbf{(D)} Simulation results showing contact-adaptive finger motion with circular objects at different contact positions (i--iii) and full finger closure without object contact (iv).}
\label{fig:2}
\end{figure*}

The proposed robotic hand consists of five finger modules, actuators connected to each module, and a printed circuit board assembly (PCBA) (Fig.~\ref{fig:2}A). Each finger module incorporates a split-stage gear--spring mechanism that enables contact-adaptive underactuated behavior in response to object contact (Fig.~\ref{fig:2}B). Each finger comprises three phalanges---the proximal phalanx (PP), middle phalanx (MP), and distal phalanx (DP)---and three joints: the metacarpophalangeal (MCP), proximal interphalangeal (PIP), and distal interphalangeal (DIP) joints.

The split-stage mechanism comprises two spur-gear trains linked through PIP Gears 1 and 2, which rotate together, and restoring springs at the PIP and DIP joints (Fig.~\ref{fig:2}B). Motor torque enters through bevel gears to the MCP gear. In the first stage, the PP idler transmits torque from the MCP gear to PIP Gear 1; the MCP and PP gears and PIP Gear 1 rotate relative to the PP frame. In the second stage, PIP Gear 2 drives the DIP gear through the MP idler. PIP Gear 2 and the MP idler rotate relative to the MP frame, whereas the DIP gear is fixed to the DP frame. Thus torque can propagate to the distal joint even when proximal phalanx motion is blocked by contact.

\newgreen{One motor drives the MCP, PIP, and DIP joints, while torsion springs at PIP and DIP resist passive flexion. Before contact, the springs hold these joints extended; under contact, transmitted torque can overcome their restoring moments (Fig.~\ref{fig:2}C). The PIP spring acts between PIP Gears 1 and 2 across the PP and MP frames, while the DIP spring acts between the MP and DP frames (Fig.~\ref{fig:2}B). To approximate human finger motion, the maximum PIP and DIP flexion angles are 90$^\circ$ and 75$^\circ$, with total spring deflections of approximately 120$^\circ$ and 105$^\circ$ including preload. Accommodating such large angular deflections generally requires more spring coils, while sufficient restoring torque requires adequate wire thickness. Both requirements make it difficult to fit the springs within the limited joint space. In the proposed hand, the PIP and DIP torsion springs each have five coils, with wire diameters of 0.9 mm and 0.7 mm, respectively. The proposed split-stage structure provides space for these relatively large springs within slender fingers.}

Fig.~\ref{fig:2}C illustrates the contact-adaptive behavior of the proposed finger mechanism under different object-contact conditions. The lines of different thicknesses represent the base, PP, MP, and DP frames, while the gear and spring symbols represent the spur gears and two torsion springs, respectively (Fig.~\ref{fig:2}C(i)). When the finger is not in contact with an object, the actuator torque is transmitted to the MCP gear through the bevel gears and then to the PIP and DIP gears through the split-stage gear train. However, the torsion springs keep the PIP and DIP joints extended, so initially only the MCP joint rotates (Fig.~\ref{fig:2}C(ii)).

Fig.~\ref{fig:2}C(iii) shows the case where an object contacts the PP. The actuator torque initially rotates the PP about the MCP joint until its motion is blocked by the contact force. However, the MCP gear can continue rotating independently of the PP frame, transmitting torque to the PIP gear through the PP gear. When the transmitted torque exceeds the restoring torque of the PIP torsion spring, the MP flexes to wrap around the object.

Fig.~\ref{fig:2}C(iv) shows the case where an object contacts the MP. The contact forces prevent further rotation of the PP and MP, while the actuator torque continues to be transmitted through the MCP and PIP gears to the DIP gear. When the transmitted torque exceeds the restoring torque of the DIP torsion spring, the DP flexes to wrap around the object.

Thus, the joint motions vary according to the object-contact conditions. Although Fig.~\ref{fig:2}C illustrates cases where only one phalanx contacts an object at a time, the mechanism can also adapt to simultaneous contact at multiple phalanges. This behavior provides a form of mechanical intelligence, allowing the finger joints to adapt to object contact without additional actuators or tactile sensing.

Fig.~\ref{fig:2}D shows simulation results of the proposed finger mechanism interacting with circular objects at different contact positions. The finger adapts its joint motions to the contact conditions and wraps around the objects (Fig.~\ref{fig:2}D(i--iii)). In the absence of an object, the joints continue to flex until they reach their mechanical rotation limits, resulting in a fully closed finger configuration (Fig.~\ref{fig:2}D(iv)).

\newgreen{The PP and MP idler gears serve two main functions: coordinating joint flexion directions and reducing finger thickness. The PP and MP idlers adjust the rotation directions between the joint gears so that the actuator torque produces flexion at all three joints. Since the finger is designed to resemble a human finger, the lengths of the PP and MP are constrained by human finger dimensions. The idler gears transmit torque between the joint gears across these fixed distances. Without the idlers, direct transmission would require larger joint gears. By selecting appropriate tooth counts for the idler gears, the diameters of the MCP, PIP, and DIP gears can be reduced while maintaining the required joint spacing. This helps maintain a thin finger profile.}

\newgreen{The two-stage architecture allows a thin PP without sacrificing the torque ratios needed for stable pinching. Proximal contact moments generally exceed distal ones \cite{ref42,ref43}. To meet these torque requirements, a single-stage gear train could result in an oversized MCP gear, as the required joint torque increases with the distance from the joint to the fingertip. In contrast, the proposed split-stage structure uses rigidly connected PIP Gears 1 and 2 with different tooth counts, providing an additional factor in the overall gear transmission ratio. This allows the MCP gear and PIP Gear 1 in the first stage to be smaller while maintaining the required torque ratios among the MCP, PIP, and DIP joints. As a result, the proposed mechanism achieves a thin, human-like finger profile across the PP, MP, and DP.}

Another structural advantage of the proposed mechanism is that it allows fully enclosed finger covers, including at the joints. As shown in the section view in Fig.~\ref{fig:2}B, the joint components are arranged coaxially. A central shaft is surrounded by a torsion spring, a sleeve bearing, and an outer finger frame. This arrangement allows all internal finger components to be enclosed by the outer cover, helping prevent dust and debris from entering the mechanism.

\begin{figure*}[!t]
\centering
\includegraphics[width=\textwidth,height=.48\textheight,keepaspectratio]{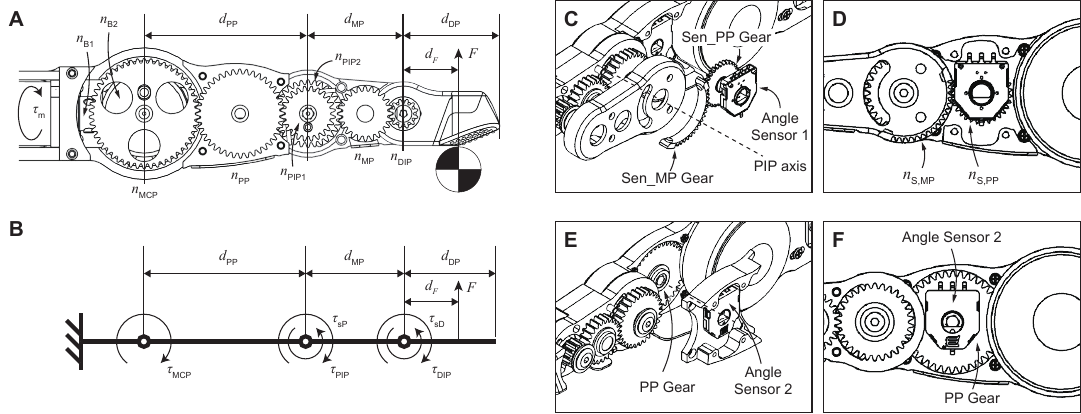}
\caption{Kinematic model of the finger module for closing-ejection avoidance and joint angle estimation. \textbf{(A)} Schematic of the split-stage gear train with an object in contact with the distal phalanx. \textbf{(B)} Simplified mechanical model of the finger. \textbf{(C)} Arrangement of Angle Sensor 1. \textbf{(D)} Angle Sensor 1 measures the PIP joint rotation through a gear pair coupled to the outer frame of the middle phalanx (MP). \textbf{(E)} Arrangement of Angle Sensor 2. \textbf{(F)} Angle Sensor 2 is coupled to the PP Gear inside the proximal phalanx (PP) frame and measures its relative rotation for joint angle estimation.}
\label{fig:3}
\end{figure*}

Compared with linkage-driven underactuated hands, the proposed gear-based mechanism offers advantages in joint rotation range and torque transmission. In linkage-driven mechanisms, joint rotation can be limited by linkage geometry and singular configurations. In contrast, the proposed mechanism allows a wide range of joint rotation. In particular, the MCP joint has no torsion spring and can rotate backward, extending the grasping range. Furthermore, while torque transmission ratios in linkage-driven mechanisms can vary with joint angles, the proposed spur gear train maintains constant transmission ratios throughout the finger's range of motion. This avoids linkage-related transmission singularities and simplifies control.

\subsection{Closing-Ejection Avoidance}

The proposed finger adapts its joint configuration in response to object contact (Fig.~\ref{fig:2}D). Its contact-adaptive behavior depends on design parameters, including gear ratios, phalanx lengths, and torsion spring properties, as well as contact force and position. Ideally, the PIP and DIP joints should remain extended when an object contacts the distal tip of the finger (Fig.~\ref{fig:2}D(i)) and flex to wrap around the object when contact shifts toward the proximal phalanx (Fig.~\ref{fig:2}D(ii)). However, an unsuitable mechanism design can cause unintended changes in joint motion under the same contact conditions. For example, if the DIP joint flexes unintentionally during fingertip pinching, the finger may slide along the object and lose contact. Such contact loss during finger closure is associated with the closing-ejection phenomenon \cite{ref42,ref43}. Therefore, in underactuated fingers, it is important to model how the design parameters affect joint torque transmission and derive appropriate conditions for avoiding closing-ejection.

To derive design conditions for avoiding closing-ejection, we analyze the torque transmission and contact forces in the proposed split-stage gear--spring mechanism. Fig.~\ref{fig:3}A shows the gear transmission structure, and Fig.~\ref{fig:3}B presents a simplified mechanical model of the finger. The analysis considers a fully extended finger with an external force $F$ acting perpendicular to the DP at a distance $d_F$ from the DIP joint. The distances between the MCP--PIP and PIP--DIP joints are denoted by $d_{\mathrm{PP}}$ and $d_{\mathrm{MP}}$, respectively, and $d_{\mathrm{DP}}$ represents the length of the DP. The contact is assumed to be frictionless, with no tangential (shear) component of the contact force. Gear and joint friction, gravity, and inertial effects are also neglected.

The actuator torque $\tau_m$ is transmitted to the MCP gear through the bevel gears. The resulting torque at the MCP joint is given by

\begin{equation}\label{eq:1}
\tau_{\mathrm{MCP}}=\tau_m\frac{n_{\mathrm{B2}}}{n_{\mathrm{B1}}}\end{equation}

Here, $n_{\mathrm{B1}}$ and $n_{\mathrm{B2}}$ are the tooth counts of the input and output bevel gears, respectively. The torques transmitted to the PIP and DIP joints are determined by the spur gear ratios:

\begin{equation}\label{eq:2}
\tau_{\mathrm{PIP}}=\tau_{\mathrm{MCP}}\frac{n_{\mathrm{PIP1}}}{n_{\mathrm{MCP}}}\end{equation}

\begin{equation}\label{eq:3}
\tau_{\mathrm{DIP}}=\tau_{\mathrm{MCP}}\frac{n_{\mathrm{PIP1}}}{n_{\mathrm{MCP}}}\frac{n_{\mathrm{DIP}}}{n_{\mathrm{PIP2}}}\end{equation}

Here, $n_{\mathrm{MCP}}$ and $n_{\mathrm{DIP}}$ are the tooth counts of the MCP and DIP gears, while $n_{\mathrm{PIP1}}$ and $n_{\mathrm{PIP2}}$ are those of PIP Gear 1 and PIP Gear 2, respectively. These equations describe the torque transmission through the two gear stages.

For the loading condition shown in Fig.~\ref{fig:3}B, moment balance about the MCP joint gives

\begin{equation}\label{eq:4}
F=\frac{\tau_{\mathrm{MCP}}}{d_{\mathrm{PP}}+d_{\mathrm{MP}}+d_F}\end{equation}

To maintain fingertip pinching without unintended joint flexion, the transmitted torques at the PIP and DIP joints must remain below the resisting moments generated by the contact force and the corresponding torsion springs. These conditions are expressed as

\begin{equation}\label{eq:5}
\tau_{\mathrm{PIP}}<F(d_{\mathrm{MP}}+d_F)+\tau_{\mathrm{sP}}\end{equation}

\begin{equation}\label{eq:6}
\tau_{\mathrm{DIP}}<Fd_F+\tau_{\mathrm{sD}}\end{equation}

Here, $\tau_{\mathrm{sP}}$ and $\tau_{\mathrm{sD}}$ denote the restoring torques of the PIP and DIP torsion springs, respectively. Equations (5) and (6) define the conditions for suppressing further joint flexion under the assumed contact condition.

Substituting Eqs.~\eqref{eq:2}--(4) into Eqs.~\eqref{eq:5} and (6) provides conditions on the contact position in terms of gear ratios, joint lengths, and spring torques. For a conservative design, the spring torques are neglected, excluding their contribution to resisting joint flexion. Under this assumption, the conditions reduce to

\begin{equation}\label{eq:7}
d_F>\frac{n_{\mathrm{PIP1}}(d_{\mathrm{PP}}+d_{\mathrm{MP}})-n_{\mathrm{MCP}}d_{\mathrm{MP}}}{n_{\mathrm{MCP}}-n_{\mathrm{PIP1}}}\end{equation}

\begin{equation}\label{eq:8}
d_F>\frac{n_{\mathrm{PIP1}}n_{\mathrm{DIP}}(d_{\mathrm{PP}}+d_{\mathrm{MP}})}{n_{\mathrm{MCP}}n_{\mathrm{PIP2}}-n_{\mathrm{PIP1}}n_{\mathrm{DIP}}}\end{equation}

These expressions require positive denominators, corresponding to $n_{\mathrm{MCP}}>n_{\mathrm{PIP1}}$ and $n_{\mathrm{MCP}}n_{\mathrm{PIP2}}>n_{\mathrm{PIP1}}n_{\mathrm{DIP}}$. Equations (7) and (8) define the lower bounds on the contact distance from the DIP joint required to suppress unintended PIP and DIP flexion, respectively, in the fully extended configuration.

To define a desired region for stable fingertip pinching, we introduce a contact-position ratio $r$, measured from the DIP joint along the DP:

\begin{equation}\label{eq:9}
rd_{\mathrm{DP}}\leq d_F\leq d_{\mathrm{DP}},\qquad 0<r\leq1\end{equation}

Here, $r$ defines the proximal boundary of the desired fingertip contact region. To satisfy both closure-suppression conditions throughout this region, the gear tooth counts and joint lengths should satisfy

\begin{equation}\label{eq:10}
\begin{aligned}\max\!\left\{0,\right.\ &\frac{n_{\mathrm{PIP1}}(d_{\mathrm{PP}}+d_{\mathrm{MP}})-n_{\mathrm{MCP}}d_{\mathrm{MP}}}{n_{\mathrm{MCP}}-n_{\mathrm{PIP1}}},\\ &\left.\frac{n_{\mathrm{PIP1}}n_{\mathrm{DIP}}(d_{\mathrm{PP}}+d_{\mathrm{MP}})}{n_{\mathrm{MCP}}n_{\mathrm{PIP2}}-n_{\mathrm{PIP1}}n_{\mathrm{DIP}}}\right\}<rd_{\mathrm{DP}}\end{aligned}\end{equation}

Equation (10) provides a design criterion for selecting the gear ratios and joint dimensions. It also allows the limiting contact-position ratio $r_{\mathrm{min}}$ to be determined from the selected design parameters. For any $r>r_{\mathrm{min}}$, the two closure-suppression conditions are satisfied throughout the specified distal contact region under the assumed loading conditions.

For the proposed finger, the selected gear tooth counts are $n_{\mathrm{MCP}}$ = 50, $n_{\mathrm{PIP1}}$ = 24, $n_{\mathrm{PIP2}}$ = 30, and $n_{\mathrm{DIP}}$ = 12. The joint dimensions are $d_{\mathrm{PP}}$ = 38.5 mm, $d_{\mathrm{MP}}$ = 22.5 mm, and $d_{\mathrm{DP}}$ = 22.6 mm. Substituting these values into Eqs.~\eqref{eq:7} and (8) gives minimum contact distances of approximately 13.04 mm and 14.50 mm, respectively. The DIP condition sets a larger minimum contact distance than the PIP condition, giving a limiting contact-position ratio $r_{\mathrm{min}}$ of approximately 0.6414. Therefore, approximately 35.9\% (8.10 mm) of the distal phalanx length is theoretically available for fingertip pinching without unintended joint flexion under the simplified model, excluding the limiting boundary.

This estimate is conservative because the model neglects the restoring spring torques and joint friction, which can resist the onset of PIP and DIP flexion under the considered loading conditions. These effects may allow a larger fingertip pinching region in practice.

The tooth counts of the PP and MP idler gears can then be determined from the joint spacing. Assuming external spur gears with a common module $m$ and collinear gear centers, their tooth counts are given by

\begin{equation}\label{eq:11}
n_{\mathrm{PP}}=\frac{d_{\mathrm{PP}}}{m}-\frac{n_{\mathrm{MCP}}+n_{\mathrm{PIP1}}}{2}\end{equation}

\begin{equation}\label{eq:12}
n_{\mathrm{MP}}=\frac{d_{\mathrm{MP}}}{m}-\frac{n_{\mathrm{PIP2}}+n_{\mathrm{DIP}}}{2}\end{equation}

Here, $n_{\mathrm{PP}}$ and $n_{\mathrm{MP}}$ represent the tooth counts of the PP and MP idler gears, respectively. The gear module and tooth counts must be selected together to satisfy the joint spacing and gear-meshing requirements.

These relations provide design conditions for suppressing unintended joint flexion associated with closing-ejection. Actual grasping behavior may also depend on contact geometry, fingertip and object materials, contact force direction, friction, and changes in finger configuration.

\subsection{Joint-Angle Estimation}

The proposed finger is driven by a single actuator, and its motion can be controlled using the actuator's motor encoder. However, the motor rotation alone does not provide the individual angles of the three finger joints. Measuring these joint angles may be useful for monitoring grasp states, detecting grasp failures, and applying LfD. A straightforward approach would be to place an angle sensor at each joint, requiring four sensors in total, including the motor encoder. However, the limited internal space of slender, human-like fingers makes it difficult to accommodate three joint sensors.

The proposed split-stage gear mechanism offers an alternative approach in which the three joint angles can be estimated using the motor encoder and two additional angle sensors located in the PP frame (Fig.~\ref{fig:3}C--F). This arrangement uses the existing gear transmission to obtain joint motion information without requiring sensors at the individual joints.

The first angle sensor (Angle Sensor 1) can be mounted on the PP frame (Fig.~\ref{fig:3}C, D). A sensor gear (Sen PP Gear) connected to its shaft meshes with another gear (Sen MP Gear) fixed to the MP frame. As the PIP joint rotates, the relative rotation between the PP and MP frames is transmitted to Angle Sensor 1 through this gear pair. Thus, the PIP joint angle can be determined from the sensor measurement and the corresponding gear ratio.

The second angle sensor (Angle Sensor 2) can be positioned further inside the PP frame (Fig.~\ref{fig:3}E, F). It is coupled to the PP Gear in the internal gear train and measures the rotation of this gear relative to the PP frame. Unlike Angle Sensor 1, which directly reflects PIP joint rotation through a dedicated gear pair, Angle Sensor 2 provides information about the relative motion within the split-stage gear transmission.

The third measurement is provided by the actuator's motor encoder. The measured motor rotation is not identical to the MCP joint rotation, as the actuator motion is coupled to the movements of all three joints through the gear mechanism.

By combining these three measurements with the kinematic relationships of the split-stage gear mechanism, the MCP, PIP, and DIP joint angles can be calculated.

Fig.~\ref{fig:3}C and D show the gear arrangement for Angle Sensor 1, while Fig.~\ref{fig:3}E and F show the arrangement for Angle Sensor 2. For the following analysis, $\theta_{\mathrm{MCP}}$, $\theta_{\mathrm{PIP}}$, and $\theta_{\mathrm{DIP}}$ denote the relative joint angles, $\theta_m$ denotes the motor rotation, and $\theta_{\mathrm{PP}}$ denotes the PP Gear rotation relative to the PP frame. All angles are defined relative to an initial reference configuration, with their positive directions selected to be consistent with the gear transmission.

The motor rotation can be expressed in terms of the three joint angles as

\begin{equation}\label{eq:13}
\begin{aligned}\theta_m=\frac{n_{\mathrm{B2}}}{n_{\mathrm{B1}}}\bigg[\theta_{\mathrm{MCP}}&+\frac{n_{\mathrm{PIP1}}}{n_{\mathrm{MCP}}}\theta_{\mathrm{PIP}}\\&+\frac{n_{\mathrm{PIP1}}}{n_{\mathrm{MCP}}}\frac{n_{\mathrm{DIP}}}{n_{\mathrm{PIP2}}}\theta_{\mathrm{DIP}}\bigg]\end{aligned}\end{equation}

Here, $n_{\mathrm{B1}}$ and $n_{\mathrm{B2}}$ are the tooth counts of the input and output bevel gears, respectively. The remaining tooth counts correspond to the spur gears defined in the preceding section. Eq.~\eqref{eq:13} shows that the motor rotation depends on the combined motion of the three joints rather than the MCP rotation alone.

The rotation of the PP Gear relative to the PP frame is related to the PIP and DIP joint angles by

\begin{equation}\label{eq:14}
\theta_{\mathrm{PP}}=\frac{n_{\mathrm{PIP1}}}{n_{\mathrm{PP}}}\theta_{\mathrm{PIP}}+\frac{n_{\mathrm{PIP1}}n_{\mathrm{DIP}}}{n_{\mathrm{PP}}n_{\mathrm{PIP2}}}\theta_{\mathrm{DIP}}\end{equation}

Here, $n_{\mathrm{PP}}$ denotes the tooth count of the PP Gear. This relationship allows Angle Sensor 2 to provide information about the combined PIP and DIP motion.

Angle Sensor 1 can measure the relative rotation between the PP and MP frames through the Sen PP Gear and Sen MP Gear (Fig.~\ref{fig:3}C, D). The angular conversion ratio of this gear pair is defined as

\begin{equation}\label{eq:15}
k_S=\frac{n_{\mathrm{S,PP}}}{n_{\mathrm{S,MP}}}\end{equation}

where $n_{\mathrm{S,PP}}$ and $n_{\mathrm{S,MP}}$ denote the tooth counts of the Sen PP Gear and Sen MP Gear, respectively. Let $\theta_{\mathrm{S1}}$ and $\theta_{\mathrm{S2}}$ denote the polarity-corrected angular measurements of Angle Sensor 1 and Angle Sensor 2, respectively. The relationships between these measurements and the corresponding joint or gear rotations are

\begin{equation}\label{eq:16}
\theta_{\mathrm{PIP}}=k_S\theta_{\mathrm{S1}},\qquad\theta_{\mathrm{PP}}=\theta_{\mathrm{S2}}\end{equation}

Substituting Eq.~\eqref{eq:16} into Eqs.~\eqref{eq:13} and~\eqref{eq:14} allows the three joint angles to be calculated directly from the angular measurements:

\begin{equation}\label{eq:17}
\theta_{\mathrm{MCP}}=\frac{n_{\mathrm{B1}}}{n_{\mathrm{B2}}}\theta_m-\frac{n_{\mathrm{PP}}}{n_{\mathrm{MCP}}}\theta_{\mathrm{S2}}\end{equation}

\begin{equation}\label{eq:18}
\theta_{\mathrm{PIP}}=k_S\theta_{\mathrm{S1}}\end{equation}

\begin{equation}\label{eq:19}
\theta_{\mathrm{DIP}}=-\frac{n_{\mathrm{PIP2}}}{n_{\mathrm{DIP}}}k_S\theta_{\mathrm{S1}}+\frac{n_{\mathrm{PP}}n_{\mathrm{PIP2}}}{n_{\mathrm{PIP1}}n_{\mathrm{DIP}}}\theta_{\mathrm{S2}}\end{equation}

These relationships can be expressed in matrix form as

\begin{equation}\label{eq:20}
\boldsymbol\theta_{\mathrm{joint}}=\mathbf A\boldsymbol\theta_{\mathrm{sensor}}\end{equation}

where

\[\boldsymbol\theta_{\mathrm{joint}}=[\theta_{\mathrm{MCP}},\theta_{\mathrm{PIP}},\theta_{\mathrm{DIP}}]^{\mathsf T},\]

\[\boldsymbol\theta_{\mathrm{sensor}}=[\theta_m,\theta_{\mathrm{S1}},\theta_{\mathrm{S2}}]^{\mathsf T}.\]

\[\mathbf A=\begin{bmatrix}\frac{n_{\mathrm{B1}}}{n_{\mathrm{B2}}}&0&-\frac{n_{\mathrm{PP}}}{n_{\mathrm{MCP}}}\\0&k_S&0\\0&-\frac{n_{\mathrm{PIP2}}}{n_{\mathrm{DIP}}}k_S&\frac{n_{\mathrm{PP}}n_{\mathrm{PIP2}}}{n_{\mathrm{PIP1}}n_{\mathrm{DIP}}}\end{bmatrix}.\]

The determinant of $\mathbf A$ is

\[\det(\mathbf A)=\frac{n_{\mathrm{B1}}}{n_{\mathrm{B2}}}k_S\frac{n_{\mathrm{PP}}}{n_{\mathrm{PIP1}}}\frac{n_{\mathrm{PIP2}}}{n_{\mathrm{DIP}}}.\]

Since all gear tooth counts are positive, the determinant is nonzero. Therefore, the three joint angles can be uniquely determined from the three angular measurements within the ideal kinematic model.

Before applying these relationships, the sensor measurements should be referenced to their initial values and assigned the appropriate rotation directions. Angular unwrapping may also be required depending on the sensor output range. This formulation provides a method for estimating all three joint angles using the motor encoder and two additional angle sensors, without placing sensors directly at each joint. The analysis assumes ideal gear transmission, and the effects of measurement errors, backlash, and structural compliance are not considered.

\subsection{Arrangement of Finger Modules in the Five-Fingered Robotic Hand}

\begin{figure}[!t]
\centering
\includegraphics[width=\columnwidth,height=.47\textheight,keepaspectratio]{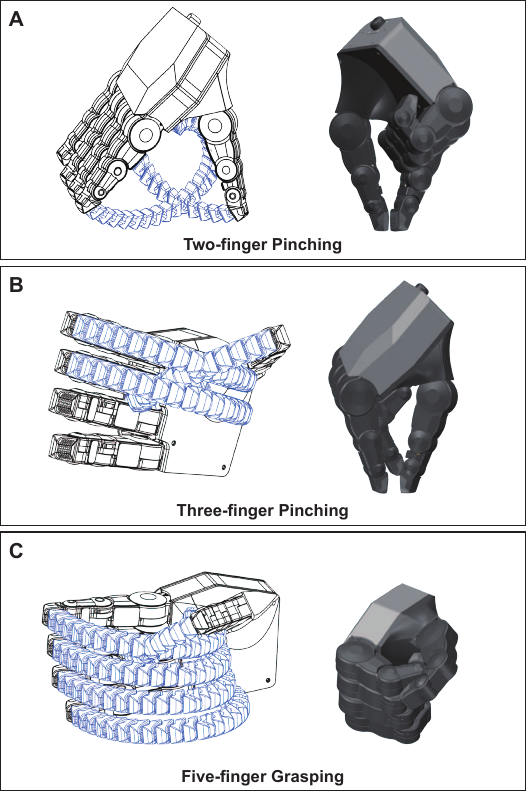}
\caption{Finger module arrangement and representative grasping modes of the five-fingered robotic hand. The five finger modules and fingertip geometries were designed to enable the following grasping modes. \textbf{(A)} Two-finger pinching. \textbf{(B)} Three-finger pinching. \textbf{(C)} Five-finger grasping.}
\label{fig:4}
\end{figure}

The five-fingered anthropomorphic robotic hand was constructed using five of the proposed finger modules. To determine their arrangement, three representative grasping motions were considered: two-finger pinching (Fig.~\ref{fig:4}A), three-finger pinching (Fig.~\ref{fig:4}B), and five-finger grasping (Fig.~\ref{fig:4}C). The positions and orientations of the finger modules within the palm were determined to enable these motions while avoiding interference between adjacent modules and with the PCBA.

In particular, the relative position and orientation of the thumb and index finger were designed to enable opposing fingertip contact during two-finger pinching while also accommodating three-finger pinching. Accordingly, the position vector $\mathbf p_{\mathrm{T/I}}$ and orientation matrix $\mathbf R_{\mathrm{T/I}}$ of the thumb MCP frame, both expressed in the index MCP frame, were specified as follows:

\[\mathbf p_{\mathrm{T/I}}=[-46.62,-11.76,-46.35]^{\mathsf T}\ \mathrm{mm},\]

\[\mathbf R_{\mathrm{T/I}}=\begin{bmatrix}-0.385&0.527&-0.758\\-0.565&-0.784&-0.258\\-0.730&0.329&0.599\end{bmatrix}.\]

The fingertip geometry was also designed to accommodate the contact configurations required for the three representative grasping motions. The total mass of the developed robotic hand was measured to be 794 g.

\subsection{Branch-Shadowing Interface and Motion Glove}

\begin{figure*}[!t]
\centering
\includegraphics[width=\textwidth,height=.48\textheight,keepaspectratio]{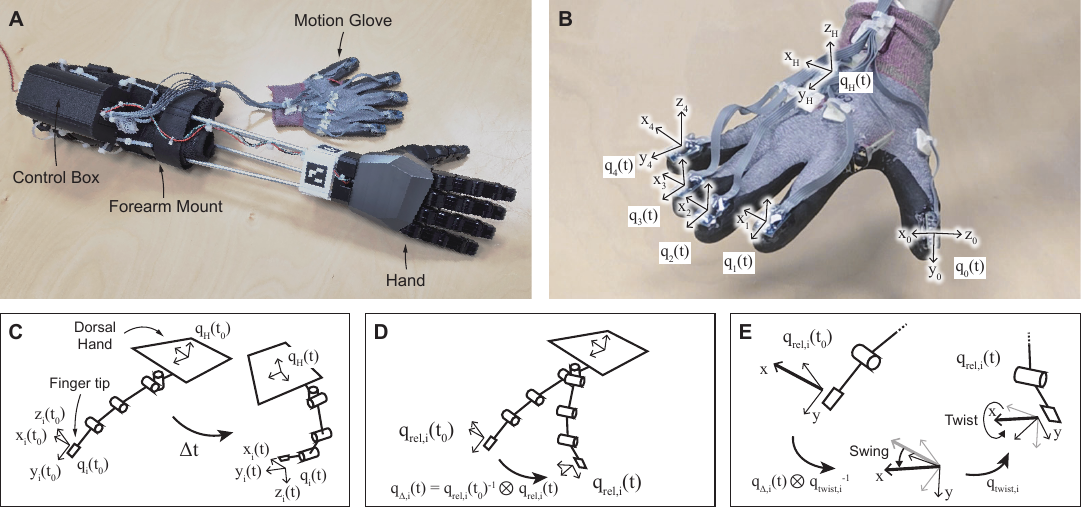}
\caption{Branch-shadowing interface and motion glove for robotic hand control. \textbf{(A)} Branch-shadowing interface. Robotic hands are mounted alongside the operator's hands using forearm mounts and controlled through motion gloves, with the aim of enhancing the operator's sense of embodiment and facilitating intuitive manipulation. \textbf{(B)} Motion glove configuration. Six attitude and heading reference system (AHRS) sensors attached to the five distal phalanges (DPs) and the dorsum of the hand provide orientation quaternions. \textbf{(C)} Changes in finger and hand sensor orientations during manipulation, including common hand rotation. \textbf{(D)} Relative quaternion calculation to remove the common hand rotation. \textbf{(E)} Swing--twist decomposition about the selected flexion axis to extract the DP rotation component for robotic finger control.}
\label{fig:5}
\end{figure*}

\subsubsection{Branch-Shadowing Interface}

\newgreen{Recent advances in artificial intelligence have improved robotic perception and decision-making \cite{ref53,ref54,ref55}. In parallel, robotic hands with many active degrees of freedom and rich sensing capabilities have been developed. Although these features offer advantages, placing more actuators and sensors within the limited space of a human-like hand can involve trade-offs in force output, durability, and compactness. This raises the question of how much hand actuation and sensing are needed when a robotic hand is combined with sufficiently advanced perception, planning, and bimanual coordination capabilities. For example, two five-DOF hands with contact-adaptive mechanisms, combined with two seven-DOF arms, provide a total of 24 active degrees of freedom, which may be sufficient for complex manipulation tasks when used with appropriate manipulation strategies.}

\newgreen{To examine this possibility, we introduce a branch-shadowing interface designed to support the operator's sense of embodiment. The branch-shadowing interface mounts each robotic hand beside the operator's hand on a forearm support (Fig.~\ref{fig:5}A). The interface uses motion gloves for finger control while the human operator provides arm movements, bimanual coordination, perception, and decision-making. This arrangement reduces dependence on other robotic subsystems, allowing us to assess the proposed hand's capabilities and limitations more clearly and explore its potential when combined with advanced perception and decision-making in future autonomous robotic systems.}

\subsubsection{Finger Motion Estimation from Relative Quaternions}

The motion gloves capture the operator's finger movements to generate control inputs for the robotic hands. Each glove incorporates six AHRS sensors: one attached to the DP of each finger and one mounted on the dorsum of the hand (Fig.~\ref{fig:5}B). Each sensor provides its orientation as a quaternion relative to a common global coordinate frame. Since each finger module of the proposed robotic hand is driven by a single actuator, the motion glove needs to extract one rotation angle from the multi-degree-of-freedom motion of each human finger.

To extract the finger rotation independently of the overall hand orientation, we first calculate the relative orientation between each finger sensor and the hand reference sensor. As shown in Fig.~\ref{fig:5}B, the five finger sensors are indexed by i = 0, \ldots{}, 4, corresponding to the thumb, index, middle, ring, and little fingers, respectively, while H denotes the hand reference sensor. Their orientation quaternions are denoted by $q_i(t)$ and $q_{\mathrm{H}}(t)$, respectively. The relative orientation of each finger sensor is calculated as

\begin{equation}\label{eq:21}
q_{\mathrm{rel},i}(t)=q_{\mathrm{H}}(t)^{-1}\otimes q_i(t),\quad i=0,\ldots,4\end{equation}

Here, $\otimes$ denotes quaternion multiplication, and the scalar-first convention is used. By expressing the finger orientation relative to the hand reference, this operation removes the common rotation of the hand and finger sensors. Fig.~\ref{fig:5}C illustrates how both the finger and hand sensors rotate during manipulation, while Fig.~\ref{fig:5}D shows how their common rotation is removed by calculating the relative orientation. To establish the initial reference configuration, $q_{\mathrm{rel},i}(t_0)$ is recorded at the initial posture. The subsequent change in finger orientation is then expressed as

\begin{equation}\label{eq:22}
q_{\Delta,i}(t)=q_{\mathrm{rel},i}(t_0)^{-1}\otimes q_{\mathrm{rel},i}(t)\end{equation}

The resulting quaternion represents the finger orientation change relative to its initial configuration, with $q_{\Delta,i}(t_0)$ equal to the identity quaternion.

However, the relative orientation change still includes multiple rotational components, such as finger flexion and abduction/adduction. Since the proposed robotic finger is controlled by a single actuator, only the rotation component associated with finger flexion is required. To extract this component, we apply quaternion swing--twist decomposition to $q_{\Delta,i}(t)$, as illustrated in Fig.~\ref{fig:5}E \cite{ref56,ref57}.

The normalized quaternion is expressed as $q_{\Delta,i}/\lVert q_{\Delta,i}\rVert$ = ($w_i$, $\mathbf{v}_i$), where $w_i$ is the scalar component and $\mathbf{v}_i=(v_{x,i},v_{y,i},v_{z,i})$ is the vector component. Under the decomposition $q_{\Delta,i}/\lVert q_{\Delta,i}\rVert$ = $q_{\mathrm{swing},i}\otimes q_{\mathrm{twist},i}$, the twist component about a prescribed unit axis $\mathbf{a}_i$ is obtained by projecting the quaternion vector component onto that axis:

\begin{equation}\label{eq:23}
q_{\mathrm{twist},i}=\frac{\left(w_i,(\mathbf v_i\cdot\mathbf a_i)\mathbf a_i\right)}{\sqrt{w_i^2+(\mathbf v_i\cdot\mathbf a_i)^2}}\end{equation}

For the proposed motion glove, each finger sensor is mounted on the DP with its local X-axis aligned with the distal joint's flexion axis at the initial reference posture. Accordingly, $\mathbf{a}_i=(1,0,0)$ in the initial finger sensor frame, and Eq.~\eqref{eq:23} reduces to

\begin{equation}\label{eq:24}
q_{\mathrm{twist},i}=\frac{(w_i,v_{x,i},0,0)}{\sqrt{w_i^2+v_{x,i}^2}}\end{equation}

The corresponding signed rotation angle is calculated as

\begin{equation}\label{eq:25}
\theta_i(t)=W_\pi\!\left[2\operatorname{atan2}(v_{x,i}(t),w_i(t))\right]\end{equation}

Here, atan2 denotes the two-argument arctangent function, and $W_\pi$ wraps the angle into the interval $[-\pi,\pi]$. This operation extracts the twist component about the selected finger axis while separating the remaining rotational components into the swing component.

Because the wrapped angle may exhibit discontinuities when crossing $\pm\pi$, the angular difference between consecutive samples is accumulated to obtain a continuous rotation angle:

\[\Theta_i[0]=\theta_i[0].\]

\begin{equation}\label{eq:26}
\Theta_i[k]=\Theta_i[k-1]+W_\pi\!\left[\theta_i[k]-\theta_i[k-1]\right],\quad k\geq1\end{equation}

Here, k denotes the sample index, and $\Theta_i[k]$ is the unwrapped finger rotation angle. This procedure removes discontinuities caused by angle wrapping, provided that the twist decomposition remains nonsingular and the rotation change between consecutive samples is less than $\pi$.

Through these operations, the motion glove extracts five rotation angles from each hand, representing the flexion-related components of the distal phalanx movements relative to the hand. These angles provide control inputs for the five single-actuator finger modules, allowing the robotic hand to follow the operator's intended finger movements without reproducing every anatomical joint motion.

\FloatBarrier
\section{Results}

\subsection{Performance Evaluation of the Finger Module}

\begin{figure*}[!t]
\centering
\includegraphics[width=\textwidth,height=.59\textheight,keepaspectratio]{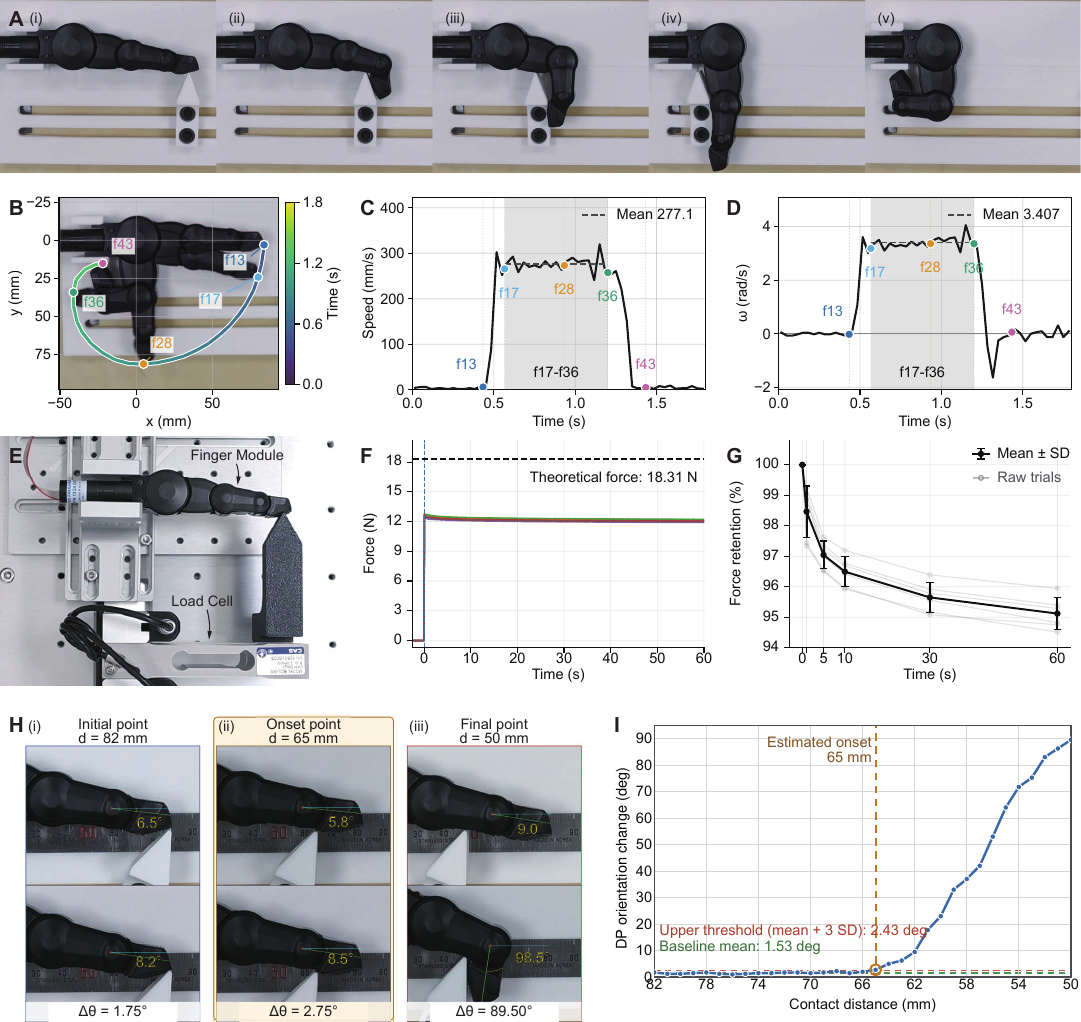}
\caption{Performance evaluation of the proposed finger module. \textbf{(A)} Contact-adaptive behavior under different contact positions and object geometries. \textbf{(B)} Fingertip trajectory and range of motion. \textbf{(C)} Fingertip speed, with a mean of 277.1 mm/s during the maximum-speed operation interval. \textbf{(D)} Fingertip angular velocity about the MCP joint center, with a mean of 3.407 rad/s during the maximum-speed operation interval. \textbf{(E)} Experimental setup for fingertip force measurement. \textbf{(F)} Fingertip force measured over 60 s, with a mean of 12.16 N, compared with the theoretical force of 18.31 N calculated without mechanical losses. \textbf{(G)} Change in fingertip force during continuous operation, showing a 4.88\% reduction after 60 s. \textbf{(H)} Experimental setup for evaluating distal phalanx orientation change associated with closing-ejection by progressively shifting the contact position from the fingertip toward the proximal phalanx. The upper and lower images show the finger before and after motor actuation, respectively. \textbf{(I)} Change in distal phalanx (DP) orientation with contact position, showing that the angular change remains small near the fingertip and increases as contact shifts proximally.}
\label{fig:6}
\end{figure*}

\begin{figure*}[!t]
\centering
\includegraphics[width=\textwidth,height=.48\textheight,keepaspectratio]{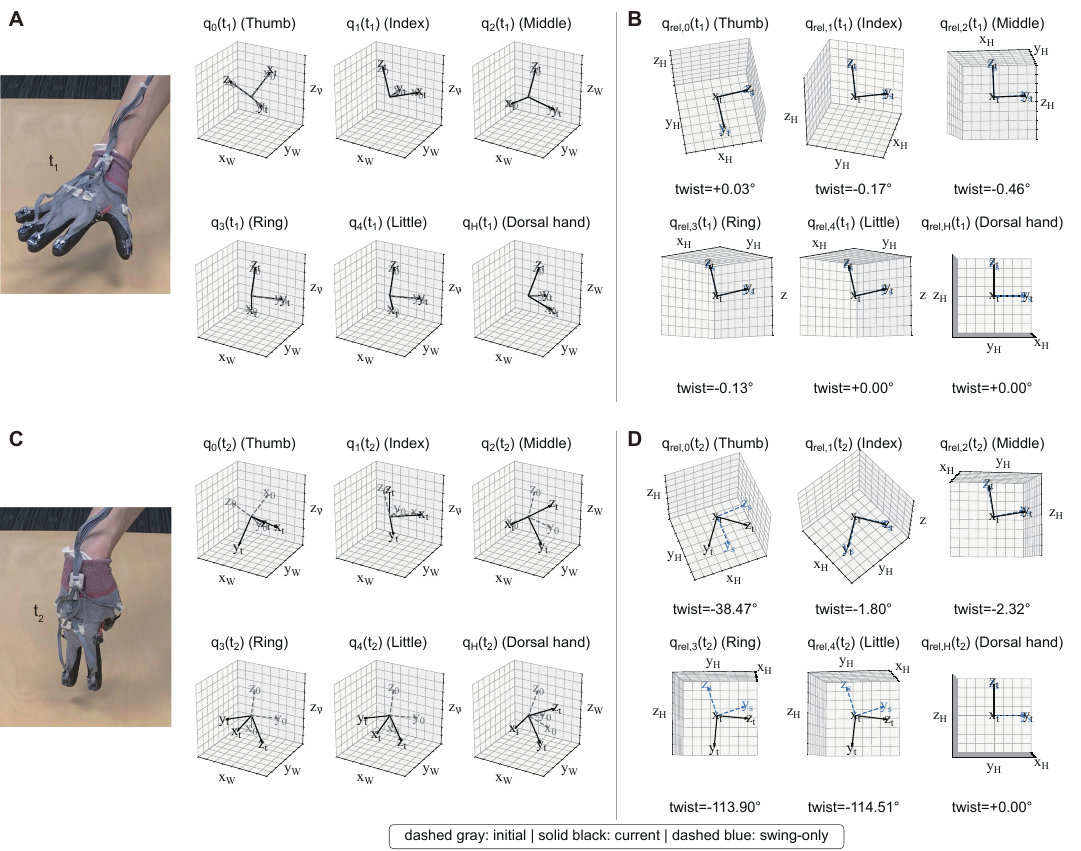}
\caption{Finger motion estimation results using the motion glove. Two representative hand postures are presented. \textbf{(A)} Initial hand posture and orientations of the six AHRS sensors in the global coordinate frame. \textbf{(B)} Relative orientations of the finger sensors with respect to the hand reference sensor at the initial posture, with the corresponding twist angles. \textbf{(C)} Second hand posture and sensor orientations in the global coordinate frame, reflecting both overall hand rotation and individual finger movements. \textbf{(D)} Relative quaternion calculation to remove the common hand rotation, followed by swing--twist decomposition to extract the flexion-related rotation components required for robotic finger control.}
\label{fig:7}
\end{figure*}

We evaluated the finger module's contact-adaptive behavior, fingertip motion, force output, and distal phalanx (DP) orientation changes associated with closing-ejection (Fig.~\ref{fig:6}). Complete-hand manipulation is demonstrated in subsequent sections.

Contact adaptation was examined using the same test object at different positions (Fig.~\ref{fig:6}A). At fingertip contact, the finger remained extended and applied force without unintended joint flexion, maintaining a fingertip pinch (Fig.~\ref{fig:6}A(i)). Contact at the MP induced DIP flexion around the object (Fig.~\ref{fig:6}A(ii)), while PP contact induced both PIP and DIP flexion (Fig.~\ref{fig:6}A(iii)). Contact beneath the finger base produced an enveloping grasp (Fig.~\ref{fig:6}A(iv)). Without contact, the finger flexed to its joint limits (Fig.~\ref{fig:6}A(v)). The split-stage gear--spring mechanism thus adapts joint configurations to contact position with one actuator, without independent joint control. The same contact-adaptive behavior is further examined with objects of different sizes, shapes, and materials in the following manipulation demonstrations.

During closure from a fully extended configuration, the fingertip followed a spiral-like inward trajectory (Fig.~\ref{fig:6}B). The phalanx lengths follow human-like proportions, allowing the fingertip to reach configurations suited to both large and small objects. Together with the contact-adaptive behavior in Fig.~\ref{fig:6}A, the spiral-like motion lets the finger adjust its configuration to the object size and contact position.

Fingertip speed and angular velocity about the MCP center were measured (Fig.~\ref{fig:6}C, D). During the maximum-speed interval, mean fingertip speed was 277.1 mm/s (Fig.~\ref{fig:6}C), and mean angular velocity about the MCP center was 3.407 rad/s (Fig.~\ref{fig:6}D). Angular velocity briefly became negative near the end of closure, indicating a reversal in the fingertip trajectory's angular direction about the MCP center near full flexion. Because this is the angular motion of the fingertip about the MCP center, it does not necessarily indicate a reversal of any individual joint rotation.

\newgreen{Fingertip force was measured for 60 s under rated-voltage stall conditions (six trials; Fig.~\ref{fig:6}E--G). The time-averaged mean was 12.16 N, compared with an ideal frictionless estimate of 18.31 N. Force declined rapidly at first, then more gradually. After 60 s, 95.12\% of the initial force remained (4.88\% loss). Heating and transmission losses may contribute, but their individual effects were not measured \cite{ref58,ref59}.}

We measured DP orientation change versus contact position to examine motion associated with closing-ejection (Fig.~\ref{fig:6}H, I). Closing-ejection refers to loss of stable contact during finger closure, potentially causing slip or ejection \cite{ref42,ref43}. A test object provided a single contact point under simplified contact conditions. The contact position shifted progressively from the fingertip toward the proximal phalanx, and DP orientation change before and after motor actuation was measured to identify additional distal rotation. Fig.~\ref{fig:6}H shows three representative contact positions and their measured changes in DP orientation. At a distance of 82 mm from the MCP center, DP orientation changed by 1.75$^\circ$. At 65 mm, it changed by 2.75$^\circ$; at 50 mm, the change reached 89.50$^\circ$.

Fig.~\ref{fig:6}I plots DP orientation change against contact distance from the MCP center. To estimate the onset of additional DP rotation, the stable region before the noticeable increase in angular change was selected by visual inspection. For 16 measurements at 82--67 mm, the baseline mean was 1.531$^\circ$ (sample standard deviation: 0.301$^\circ$). The mean plus three standard deviations gave a threshold of 2.434$^\circ$. At 65 mm, DP orientation change first exceeded this threshold (2.75$^\circ$). This was taken as the estimated onset of additional DP rotation; the change increased at more proximal contacts.

The theoretical boundary for suppressing DIP flexion was 75.50 mm from the MCP center. The estimated onset of additional DP rotation occurred 10.5 mm closer to the MCP center. The restoring spring torques and joint friction neglected in the theoretical model may contribute to this difference. Although the experiment measured DP orientation rather than the relative DIP joint angle and did not directly demonstrate object ejection or grasp failure, it provides evidence of contact-position-dependent DP motion associated with closing-ejection.

\subsection{Evaluation of Finger Motion Estimation Using the Motion Glove}

We evaluated motion-glove finger motion estimation using two hand postures (Fig.~\ref{fig:7}). The initial posture defined the reference configuration (Fig.~\ref{fig:7}A), allowing the subsequent motion to be compared with a known baseline. In the second, the wrist bent downward in a V-sign gesture, with the index and middle fingers extended and the others flexed (Fig.~\ref{fig:7}C). These postures test how relative quaternions and swing--twist decomposition separate finger movements from hand rotation.

At the initial posture, the six AHRS sensors reported orientations in the global frame (Fig.~\ref{fig:7}A), containing both hand rotation and relative finger orientations. Calculating relative quaternions with respect to the dorsal hand sensor removed the rotation common to hand and finger sensors (Fig.~\ref{fig:7}B). The extracted twist angles were close to zero for all five fingers, ranging from $-$0.46$^\circ$ to +0.03$^\circ$, providing a reference for subsequent movements.

At the second posture (Fig.~\ref{fig:7}C), global sensor orientations combined hand rotation and individual finger motion. After removing common hand rotation, applying swing--twist decomposition extracted the flexion-related components from each finger sensor's relative orientation change (Fig.~\ref{fig:7}D).

Twist angles were $-1.80^\circ$ and $-2.32^\circ$ for the index and middle fingers, indicating that they remained nearly extended. In contrast, the flexed ring and little fingers had angles of $-113.90^\circ$ and $-114.51^\circ$, respectively. The thumb showed an intermediate angle of $-38.47^\circ$, reflecting its different posture. Thus, the method distinguished finger flexion-related motion despite changes in overall hand orientation.

The extracted angles represent distal phalanx rotation about selected flexion axes, not individual anatomical joint angles. Nevertheless, they provide the rotation information needed to control single-actuator finger modules. The method thus translates multi-DOF human finger motion into a reduced set of robotic commands.

\subsection{Adaptive Grasping and Dexterous Manipulation with the Five-Fingered Robotic Hand}

\begin{figure*}[!t]
\centering
\includegraphics[width=\textwidth]{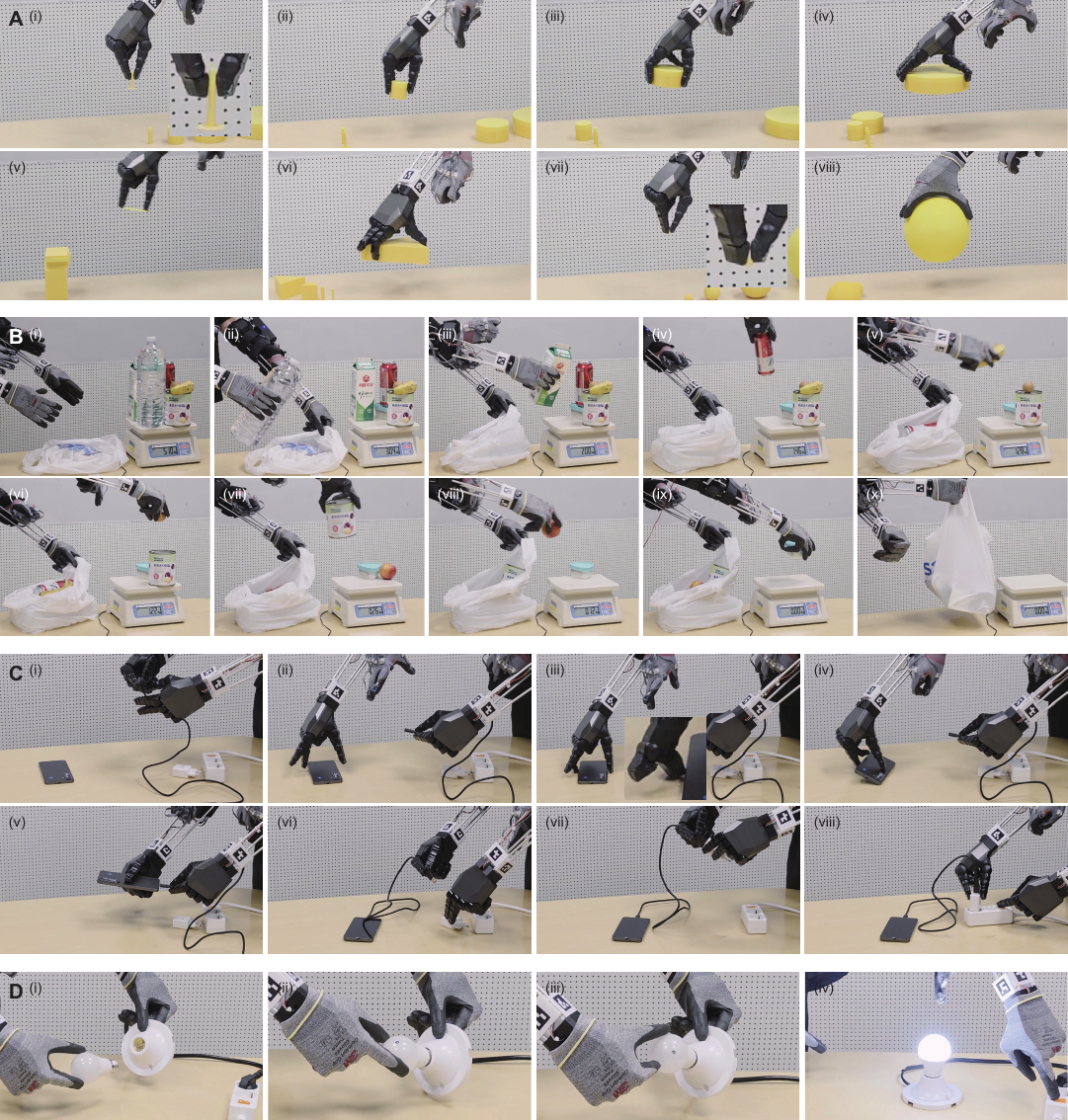}
\caption{Adaptive grasping and dexterous manipulation with the proposed five-fingered robotic hand. \textbf{(A)} Grasping objects of various sizes and geometries. (i) Two-finger pinching of a 5 mm diameter cylinder. (ii) Three-finger pinching of a 40 mm diameter cylinder. (iii) Four-finger grasping of an 80 mm diameter cylinder. (iv) Five-finger grasping of a 165 mm diameter cylinder. (v) Grasping a thin plate 3 mm thick. (vi) Grasping a large triangular prism 165 mm wide. (vii) Pinching a small sphere 5 mm in diameter. (viii) Grasping a large sphere 165 mm in diameter with the robotic hand wearing a glove. \textbf{(B)} Grasping and packing everyday grocery items into a plastic bag. (i--ii) Handling a plastic bag and a 2 L water bottle. (iii) Milk carton. (iv) Beverage can. (v) Banana. (vi) Egg. (vii) Canned fruit. (viii) Peach. (ix) Yogurt container. (x) Lifting the bag containing the packed items. \textbf{(C)} Smartphone pickup and charging cable manipulation. (i) Bimanual manipulation of a USB-C charging cable. (ii--v) Scooping up a smartphone lying screen-down on a flat surface without a protective case. (vi) Inserting the USB-C connector into the smartphone. (vii) Connecting the USB-A connector to the charger. (viii) Plugging the charger into a 220 V power strip and switching on the power. \textbf{(D)} In-hand manipulation for light bulb installation. (i) Picking up the light bulb and socket. (ii--iii) Rotating the bulb within the fingers to screw it into the socket. (iv) Completing the installation and switching on the light.}
\label{fig:8}
\end{figure*}

We demonstrated five-fingered hand manipulation across object sizes, geometries, materials, and everyday tasks using the branch-shadowing interface (Fig.~\ref{fig:8}). These experiments examined how the contact-adaptive mechanism and human-like hand configuration support such tasks with only five active degrees of freedom and no tactile sensing.

Fig.~\ref{fig:8}A(i--iv) shows cylinders measuring 5 $\times$ 40, 40 $\times$ 40, 80 $\times$ 40, and 165 $\times$ 40 mm (diameter $\times$ length), grasped using two, three, four, and five fingers, respectively. Although similar finger flexion commands were applied through the motion glove, joint configurations differed with object size and contact conditions. The changes resulted from mechanical interaction rather than explicit joint-angle commands. Rather than following predetermined joint configurations, the underactuated MCP, PIP, and DIP joints adapted mechanically through contact with each object. For the 165 mm cylinder (Fig.~\ref{fig:8}A(iv)), the proximal joints remained largely extended while the distal phalanges flexed against its surface. The arrangement and split-stage gear--spring mechanism thus enabled the hand to grasp an object substantially wider than the individual fingers, without independent control of each finger joint.

The hand pinched a 3 mm thick plate, demonstrating grasping with a small contact region (Fig.~\ref{fig:8}A(v)) and grasped a 165 mm wide triangular prism (Fig.~\ref{fig:8}A(vi)). With a simple five-finger closing posture, the fingers adapted passively to the prism's different contact surfaces. This did not require joint-angle optimization or independent control. The hand also pinched a 5 mm sphere (Fig.~\ref{fig:8}A(vii)) and grasped a 165 mm sphere (Fig.~\ref{fig:8}A(viii)). For the large sphere, the hand wore a glove to increase friction at contact surfaces. This shows that conventional hand coverings can modify surface properties without changes to the internal finger mechanism.

\newgreen{In a continuous grocery-packing task (Fig.~\ref{fig:8}B), the hands handled a flexible plastic bag, packed grocery items, and lifted the bag. The items included a 2 L water bottle (2.06 kg), milk carton (1.06 kg), beverage can (0.52 kg), banana (0.18 kg), egg (0.06 kg), canned fruit (0.96 kg), peach (0.14 kg), and yogurt container (0.12 kg). The heavy water bottle was grasped from its side using contact-adaptive finger closure. Delicate items, including the banana, egg, and peach, were handled without tactile sensing or independent joint control. Unlike rigid objects, the flexible bag required the hands to adapt as its shape changed during manipulation. These tasks demonstrate handling of objects with different geometries, stiffnesses, and surface properties.}

\newgreen{A challenging task was picking up a smartphone lying flat on a table (Fig.~\ref{fig:8}C). The sequence began with bimanual manipulation of a USB-C cable (Fig.~\ref{fig:8}C(i)). The hands then picked up a Samsung Galaxy S24+ (158.5 $\times$ 75.9 $\times$ 7.7 mm, 196 g; Fig.~\ref{fig:8}C(ii--v)). The phone lay screen-down directly on the tabletop without a protective case, making conventional fingertip pinching difficult. Instead, the hand used tabletop contact to scoop it up: fingertip contact with the table induced DIP flexion, rotating the distal phalanges to lift one edge (Fig.~\ref{fig:8}C(iii)). The hand then established a supporting grasp and lifted the phone (Fig.~\ref{fig:8}C(iv--v)). The sequence demonstrates how tabletop contact is used by the underactuated mechanism to produce a scooping motion without independently actuating the DIP joints.}

The hands then connected the smartphone to a power source (Fig.~\ref{fig:8}C(vi--viii)). They inserted the USB-C connector into the phone (Fig.~\ref{fig:8}C(vi)), the USB-A connector into the charger (Fig.~\ref{fig:8}C(vii)), and the charger into a 220 V power strip before switching on power (Fig.~\ref{fig:8}C(viii)). The flexible cable was handled without tactile sensing or independent finger-joint control. These results illustrate how human coordination and perception complement the mechanical capabilities of an underactuated hand in flexible-object manipulation and connector insertion.

\begin{figure*}[!t]
\centering
\includegraphics[width=\textwidth]{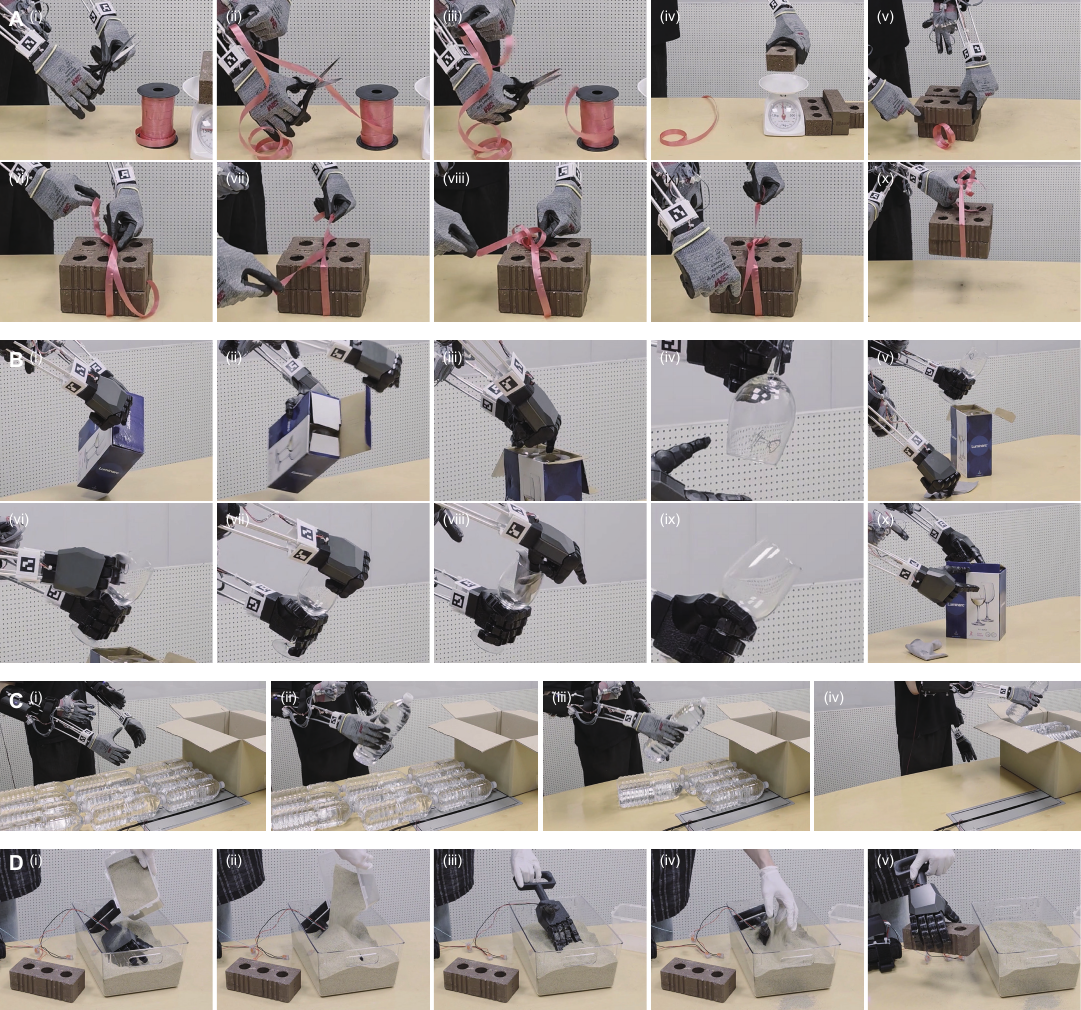}
\caption{Complex manipulation and robustness demonstrations with the proposed five-fingered robotic hand. \textbf{(A)} Cutting a rope with scissors, tying bricks together, and carrying the bundled bricks. (i) Picking up scissors and inserting the fingers into the handles in midair. (ii--iii) Cutting a rope with the scissors while held in midair. (iv--v) Rearranging and stacking bricks in different orientations. (vi--ix) Tying two knots around the bricks. (x) Lifting and carrying the bundled bricks with one hand. \textbf{(B)} Unboxing, inspection, cleaning, and repackaging of a fragile object. Taking a wine glass out of its box, cleaning its exterior and interior with a cloth, and repackaging it. (i--ii) Opening the box. (iii) Inserting slender fingers into the narrow space to lift the wine glass out of the box. (iv) Inspecting contamination on the exterior and interior surfaces of the glass. (v) Picking up a cleaning cloth with the other hand. (vi) Wiping the exterior surface. (vii) Reorienting the glass in midair through coordinated bimanual manipulation. (viii) Wiping the interior surface. (ix) Inspecting the cleaned glass. (x) Repackaging the glass in its box. \textbf{(C)} Continuous high-load operation. (i--iv) Continuously transferring twelve 2 L water bottles into a box by grasping them from their sides. \textbf{(D)} Enclosed finger structure and operation after sand exposure. (i--ii) Covering the robotic hand with sand. (iii--iv) Turning the hand over and exposing the opposite side to sand. (v) Verifying normal operation after removing the hand from the sand.}
\label{fig:9}
\end{figure*}

In-hand manipulation was examined through light bulb installation (Fig.~\ref{fig:8}D). The hands picked up a bulb and socket and held them in the air (Fig.~\ref{fig:8}D(i)). One hand held the socket while the other repeatedly rotated the bulb within its fingers to screw it in (Fig.~\ref{fig:8}D(ii--iii)); surface markings indicated its rotation. The oblique arrangement of the thumb and index finger enabled coordinated finger motions to generate a turning moment on the bulb. By repeating the rotation and adjusting the grasp while both objects were held in midair, the hand progressively installed the bulb. It was then switched on (Fig.~\ref{fig:8}D(iv)). The demonstration shows that a hand with one actuator per finger can perform in-hand rotation beyond a fixed grasp when combined with operator-controlled arm movements and bimanual coordination.

\subsection{Complex Manipulation and Robustness Demonstrations}

We further examined complex manipulation through tool use, flexible objects, fragile and slippery object handling, repeated high-load operation, and sand exposure (Fig.~\ref{fig:9}). These tasks evaluated how contact adaptation, slender finger geometry, and a reduced number of actuators contribute to operations beyond simple grasping.

Fig.~\ref{fig:9}A shows tool use, brick manipulation, and rope tying. The hand picked up scissors and inserted its fingers into the handles in midair (Fig.~\ref{fig:9}A(i)). The index and middle fingers entered one handle loop, while the ring finger flexed against the same handle from outside. Together, the index, middle, and ring fingers securely held one handle, while the thumb entered the opposite handle loop to open and close the blades. The slender finger structure allowed multiple fingers to fit within and around the narrow scissor handles, while contact adaptation accommodated the joint configurations needed to hold and operate the tool (Fig.~\ref{fig:9}A(ii--iii)).

The scissor task highlights a mismatch between operator motion and the joint configurations required under contact. During teleoperation, the operator does not interact with scissors in the same configuration as the robotic hand. Thus, even accurately measured finger movements may not produce the robotic joint angles required by object contact. The underactuated joints instead adapt mechanically to the handles. The use of fewer, larger actuators also supports sustained force generation when object contact constrains the commanded motion. Thus, the hand accommodates teleoperation and contact constraints without independent joint control.

The hands next rearranged four differently oriented bricks (Fig.~\ref{fig:9}A(iv--v)), using strategies including inserting slender fingers into brick holes and transferring bricks between hands. Thin finger geometry enabled access to the brick holes and narrow contact regions that would be difficult to reach with bulkier mechanisms. The hands then wrapped rope around the bricks and tied two knots through bimanual manipulation (Fig.~\ref{fig:9}A(vi--ix)). Although the rope is highly deformable and can take many configurations, the task was performed mainly through coordinated arm movements and grasping rather than extensive independent finger-joint manipulation. Finally, the bundled bricks were lifted and carried with one hand (Fig.~\ref{fig:9}A(x)). This sequence shows how a relatively simple robotic hand can perform multi-step manipulation through human bimanual coordination and decision-making.

Fig.~\ref{fig:9}B shows unboxing, inspecting, cleaning, and repackaging a wine glass. The robotic hands first opened the box (Fig.~\ref{fig:9}B(i--ii)) and inserted slender distal phalanges into the narrow clearance to remove the glass (Fig.~\ref{fig:9}B(iii)). After removal, the operator inspected the exterior and interior surfaces for contamination while holding the glass in midair (Fig.~\ref{fig:9}B(iv)), then used the other hand to pick up a cleaning cloth and wipe the glass exterior while holding it in midair (Fig.~\ref{fig:9}B(v--vi)).

Handling the wine glass required a secure, sustained grip on its smooth, slippery surface while avoiding excessive force on the fragile object. The relatively large actuators accommodated in the hand supplied sustained gripping force while the fragile glass was held in midair. The hands reoriented the glass through bimanual manipulation (Fig.~\ref{fig:9}B(vii)), wiped its interior (Fig.~\ref{fig:9}B(viii)), inspected it, and repackaged it (Fig.~\ref{fig:9}B(ix--x)). These operations illustrate how slender finger geometry and sustained gripping force support manipulation of a fragile, slippery object without changing the grasping mechanism.

Repeated high-load grasping was examined by transferring twelve 2 L water bottles consecutively into a box (Fig.~\ref{fig:9}C). Each was grasped from the side and transferred without interruption. This task required repeated high gripping forces and continuous actuator operation. With fewer actuators than independently driven multi-joint hands, the proposed design provides space for larger motors, supporting sustained force and potentially improving heat dissipation compared with densely integrated joint motors. Completing twelve consecutive transfers demonstrated repeated high-load operation over the tested sequence.

\newgreen{The enclosed gear--spring mechanism was tested by sand exposure (Fig.~\ref{fig:9}D). Sand and other particles can interfere with gear transmissions. Here, the outer finger frames and sleeve bearings enclose the gear train and joint components, reducing their direct exposure. The hand was completely buried in sand (Fig.~\ref{fig:9}D(i--ii)), turned over, and buried again (Fig.~\ref{fig:9}D(iii--iv)). The fingers continued to operate normally after removal from the sand (Fig.~\ref{fig:9}D(v)), suggesting that the enclosed finger structure helps protect internal transmission from particulates, although the experiment did not establish a standardized dust-ingress protection rating.}

Together, the demonstrations show that contact adaptation, slender fingers, and sustained actuator force support manipulation of tools, fragile and slippery objects, flexible materials, and heavy objects. Human perception, decision-making, and bimanual coordination provided through the branch-shadowing interface complement these features despite the limited active degrees of freedom.

\FloatBarrier
\section{Discussion}

\subsection{Teleoperation and Contact Adaptation}

Motion-glove teleoperation involves a mismatch between operator finger movements and the joint configurations needed for object grasping \cite{ref14,ref23,ref60}. Because the operator's fingers do not contact the grasped object, even accurate human joint-angle tracking and mapping to a high-DOF robotic hand cannot guarantee stable grasps. This mismatch is related to the embodiment gap in teleoperation and the need for contact-aware motion retargeting.

The motion glove conveys intended finger movements using a few rotation angles rather than replicating every human joint. The contact-adaptive mechanism accommodates differences between these commands and the joint configurations required by object contact. Fewer actuators also provide space for larger motors, helping maintain grip when object contact prevents the fingers from following commanded motion. Together, these features reduce the mismatch between object-free human motion and contact-constrained robotic motion without precise joint-angle matching or tactile sensing (Figs.~\ref{fig:8}A(vi), 8B(ix), 9A(ii), 9A(v), and 9B(ii)).

Despite having far fewer active degrees of freedom than a human hand, the proposed hand performed complex tasks by combining contact adaptation with human perception, decision-making, and bimanual coordination. The results suggest that manipulation capability is not determined solely by the number of independently actuated finger joints but also by how control inputs are translated into physical interaction with objects.

\subsection{Design Implications and Limitations}

The demonstrations show the value of slender fingers in confined spaces. Thin distal phalanges entered the narrow clearance between the wine glass and its packaging (Fig.~\ref{fig:9}B), accessed closely packed groceries (Fig.~\ref{fig:8}B), and reached brick holes during manipulation (Fig.~\ref{fig:9}A). Slender proximal and middle phalanges allowed multiple fingers to fit within and around narrow scissor handles (Fig.~\ref{fig:9}A). Across these tasks, maintaining a slender, human-like finger profile was particularly useful for accessing confined regions of objects and tools designed for human hands.

In contrast, palm thickness did not appear to be a major constraint in the demonstrated tasks, even though the proposed hand has a relatively thin palm. Future designs could retain slender fingers while allocating more palm space for larger motors with lower reduction ratios.

The thumb is fixed in orientation relative to the palm, without independent abduction/adduction. Although this limitation did not prevent the completion of the manipulation tasks demonstrated in this study, adjustable thumb orientation might help stabilize flat objects against the palm or operate tools such as joysticks. Similarly, independent control of the index finger PIP joint could facilitate operation of trigger-actuated tools. These examples suggest that some tasks may justify additional active degrees of freedom despite trade-offs in force output, durability, and hand size.

\newgreen{The hand had no fingertip tactile sensors, and demonstrations used no dedicated tactile or haptic feedback. The operator relied primarily on visual feedback to assess object interactions, while the contact-adaptive joints conformed to object surfaces even when glove commands did not precisely match the required joint configurations. These characteristics may have reduced the need for tactile feedback in the tested tasks. However, tactile sensing may be beneficial for delicate or deformable objects requiring precise force control or when vision is limited \cite{ref62,ref63,ref64}.}

\newgreen{The forearm-mounted branch-shadowing interface provides no independent wrist rotation. The operator instead achieved the required hand positions and orientations through arm and upper-body movements, but wrist articulation could improve efficiency by reducing such compensatory motions. Future work could investigate a two-DOF wrist with a load-support structure transferring hand weight to the forearm, allowing orientation changes without requiring the operator to support the full hand load at the wrist.}

The forearm-mounted hands also impose weight on the operator, potentially increasing effort during prolonged use. Possible load support includes an external cable-suspension system, similar to upper-limb gravity-compensation and teleoperation devices \cite{ref65,ref66}, or a wearable exoskeleton transferring part of the hand weight from the arms to the torso \cite{ref67}. Both approaches could reduce the physical burden of supporting the robotic hands during extended tasks while preserving mobility for bimanual coordination.

Although branch-shadowing reduces dependence on robotic arms and autonomous perception, performance still depends on the operator's skills and strategies. The results therefore assess the hand with human perception, decision-making, and bimanual coordination rather than directly predicting autonomous robotic performance. Autonomous evaluations remain necessary to establish transfer to robotic systems.

\subsection{Future Integration with Learning from Demonstration}

\begin{figure}[!t]
\centering
\includegraphics[width=\columnwidth,height=.38\textheight,keepaspectratio]{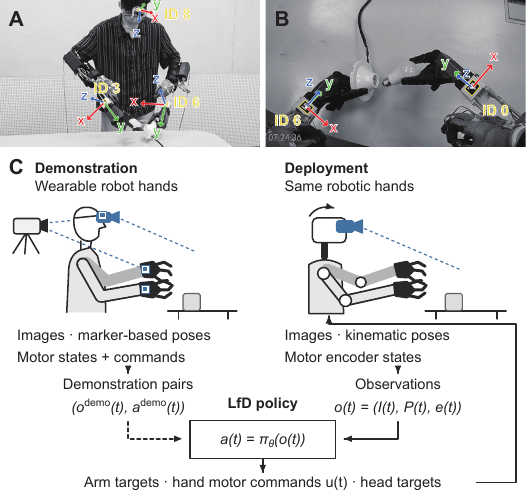}
\caption{Proposed integration of the branch-shadowing interface with learning from demonstration (LfD) for autonomous manipulation. \textbf{(A)} During demonstration, an external camera would track the positions and orientations of both robotic hands and the operator's head using visual markers. \textbf{(B)} A head-mounted camera would capture egocentric images of the task scene and estimate the poses of the robotic hands relative to the camera using visual markers. \textbf{(C)} Proposed LfD framework. Demonstrations would provide camera images, tracked head and hand poses, and finger motor states and commands for policy learning. During deployment, the same robotic hands would be mounted on a robot, with kinematically estimated poses and motor encoder states used as observations. The learned policy would generate head and arm targets and finger motor commands for autonomous manipulation.}
\label{fig:10}
\end{figure}

A possible extension of branch-shadowing is learning from demonstration (LfD) for autonomous manipulation with the same hands (Fig.~\ref{fig:10}). During demonstrations, an external camera could track both hands and the operator's head, while a head-mounted camera could capture egocentric images and estimate hand poses relative to the camera. Calibrated coordinate transformations could express head and hand poses in a common global frame, linking the external and egocentric camera observations. The finger motor encoder states and motion-glove commands could then be recorded throughout each manipulation demonstration for subsequent policy learning.

The data could form demonstration pairs $(o^{\mathrm{demo}}(t), a^{\mathrm{demo}}(t))$ (Fig.~\ref{fig:10}C). Observations $o(t)=(I(t),P(t),e(t))$ consist of camera images $I(t)$, head and hand poses $P(t)$, and motor encoder states $e(t)$. The corresponding actions would include target head poses, target hand poses for arm control, and finger motor commands derived from the motion glove. An LfD policy $\pi_\theta$ could map observations to actions, $a(t)=\pi_\theta(o(t))$, where $\theta$ denotes the learned parameters.

\newgreen{During deployment, the same hands could be mounted on a robot with a head-mounted camera. Robot kinematics would provide head and hand poses instead of marker tracking; images and encoders would supply the other observations. The policy would generate head and arm targets and finger motor commands. Using the same robotic hands for demonstration and deployment could reduce hand-level motion retargeting, although differences in arm kinematics and observation systems would still have to be addressed. Practical deployment would require robotic arms and a movable camera platform capable of reaching poses corresponding to those recorded for the operator's hands and head during demonstration.}

Another possible LfD extension is wrist force/torque (F/T) sensing. In the branch-shadowing interface, the weight of a grasped object reaches the operator through the forearm mount. This load perception may affect the operator's manipulation strategies and decisions. If this information proves useful, wrist F/T sensors could be used during both demonstration and deployment, with their measurements included among policy observations to account for object loads.

\FloatBarrier
\section{Conclusion}

\newgreen{In this study, we presented a five-fingered anthropomorphic robotic hand based on a split-stage gear--spring mechanism. Each finger uses a single actuator to achieve contact-adaptive behavior, allowing its joint configuration to adjust mechanically to object contact. The proposed architecture is designed to address several trade-offs in robotic hand design by combining contact-adaptive finger motion, slender, fully enclosed fingers, and sustained gripping force. We developed a simplified mechanical model of the underactuated finger and derived design conditions intended to suppress closing-ejection. We also introduced a motion glove based on quaternion swing--twist decomposition and a branch-shadowing interface designed to support the operator's sense of embodiment during teleoperation and evaluate the hand's manipulation capabilities with reduced dependence on other robotic subsystems.}

\newgreen{Through quantitative experiments and manipulation demonstrations, we evaluated the finger module's motion, force output, and contact-adaptive behavior, and demonstrated a range of tasks involving objects of different sizes, shapes, and materials. These included smartphone pickup, in-hand manipulation, tool use, flexible object handling, continuous high-load operation, and complex bimanual tasks. These demonstrations suggest that even a relatively simple robotic hand with limited active degrees of freedom and no tactile sensing can support complex manipulation when contact-adaptive mechanics are combined with capable perception, decision-making, and bimanual coordination. Future work could investigate whether comparable performance can be achieved in autonomous robotic systems.}

\section*{Supplementary Material}
\noindent\textbf{Supplementary Video S1.} Demonstrations of the proposed robotic hand, including the full-length video and highlight video. Available at \href{https://github.com/HGShin-robot/contact-adaptive-robotic-hand/releases/latest}{GitHub Releases}.

\section*{Author Contributions}
\begingroup\small\setlength{\parindent}{0pt}
Conceptualization: H.G.S.

Methodology: H.G.S. (robotic hand mechanism, hardware, and hand-cover design, motion glove, and branch-shadowing interface).

Software: H.G.S. (high-level control, motion-glove and interface software, and system integration); S.J. (low-level hand motor control and hand--control box communication).

Investigation: H.G.S.

Visualization: H.G.S.

Supervision: H.G.S.

Writing -- original draft: H.G.S.

Writing -- review \& editing: H.G.S.
\endgroup

\section*{Competing Interests}
\noindent The authors declare no competing interests.


\begin{thebibliography}{99}

\bibitem{ref1} K. Tai, A. R. El-Sayed, M. Shahriari, M. Biglarbegian, and S. Mahmud, ``State of the art robotic grippers and applications,'' \emph{Robotics}, vol. 5, art. no. 11, 2016.

\bibitem{ref2} M. Chitroda and D. B. Patle, ``A review on technologies in robotic gripper,'' \emph{Int. J. Adv. Eng. Nano Technol.}, vol. 10, pp. 1--5, 2023.

\bibitem{ref4} S. Sankar, W.-Y. Cheng, J. Zhang, A. Slepyan, M. M. Iskarous, R. J. Greene, R. DeBrabander, J. Chen, A. Gupta, and N. V. Thakor, ``A natural biomimetic prosthetic hand with neuromorphic tactile sensing for precise and compliant grasping,'' \emph{Sci. Adv.}, vol. 11, art. no. eadr9300, 2025. \href{https://doi.org/10.1126/sciadv.adr9300}{doi: 10.1126/sciadv.adr9300}.

\bibitem{ref5} H. Duan, P. Wang, Y. Yang, D. Li, W. Wei, Y. Luo, and G. Deng, ``Reactive human-to-robot dexterous handovers for anthropomorphic hand,'' \emph{IEEE Trans. Robot.}, vol. 41, pp. 742--761, 2025. \href{https://doi.org/10.1109/TRO.2024.3508194}{doi: 10.1109/TRO.2024.3508194}.

\bibitem{ref7} M. G. Catalano, G. Grioli, E. Farnioli, A. Serio, C. Piazza, and A. Bicchi, ``Adaptive synergies for the design and control of the Pisa/IIT SoftHand,'' \emph{Int. J. Robot. Res.}, vol. 33, pp. 768--782, 2014. \href{https://doi.org/10.1177/0278364913518998}{doi: 10.1177/0278364913518998}.

\bibitem{ref8} U. Kim, D. Jung, H. Jeong, J. Park, H.-M. Jung, J. Cheong, H. R. Choi, H. Do, and C. Park, ``Integrated linkage-driven dexterous anthropomorphic robotic hand,'' \emph{Nat. Commun.}, vol. 12, art. no. 7177, 2021. \href{https://doi.org/10.1038/s41467-021-27261-0}{doi: 10.1038/s41467-021-27261-0}.

\bibitem{ref13} C. Wang, H. Shi, W. Wang, R. Zhang, L. Fei-Fei, and C. K. Liu, ``DexCap: Scalable and portable mocap data collection system for dexterous manipulation,'' in \emph{Proc. Robotics: Science and Systems (RSS)}, 2024. \url{https://arxiv.org/abs/2403.07788}.

\bibitem{ref14} M. Xu, H. Zhang, Y. Hou, Z. Xu, L. Fan, M. Veloso, and S. Song, ``DexUMI: Using human hand as the universal manipulation interface for dexterous manipulation,'' in \emph{Proc. Mach. Learn. Res.}, vol. 305, pp. 437--459, 2025. \url{https://proceedings.mlr.press/v305/xu25b.html}.

\bibitem{ref15} T. G. W. Lum, O. Y. Lee, C. K. Liu, and J. Bohg, ``Crossing the human-robot embodiment gap with sim-to-real RL using one human demonstration,'' in \emph{Proc. Mach. Learn. Res.}, vol. 305, pp. 4418--4441, 2025. \url{https://proceedings.mlr.press/v305/lum25a.html}.

\bibitem{ref16} S. Stipeti{\'c}, D. {\v{Z}}arko, and M. Popescu, ``Ultra-fast axial and radial scaling of synchronous permanent magnet machines,'' \emph{IET Electr. Power Appl.}, vol. 10, pp. 658--666, 2016. \href{https://doi.org/10.1049/iet-epa.2016.0014}{doi: 10.1049/iet-epa.2016.0014}.

\bibitem{ref17} J. Zhu, C. Jiao, I. Dominguez, S. Yu, and H. Su, ``Design and backdrivability modeling of a portable high torque robotic knee prosthesis with intrinsic compliance for agile activities,'' \emph{IEEE/ASME Trans. Mechatron.}, vol. 27, pp. 1837--1845, 2022. \href{https://doi.org/10.1109/TMECH.2022.3176255}{doi: 10.1109/TMECH.2022.3176255}.

\bibitem{ref19} P. M. Wensing, A. Wang, S. Seok, D. Otten, J. H. Lang, and S. Kim, ``Proprioceptive actuator design in the MIT Cheetah: Impact mitigation and high-bandwidth physical interaction for dynamic legged robots,'' \emph{IEEE Trans. Robot.}, vol. 33, pp. 509--522, 2017. \href{https://doi.org/10.1109/TRO.2016.2640183}{doi: 10.1109/TRO.2016.2640183}.

\bibitem{ref21} S. Yu, T.-H. Huang, X. Yang, C. Jiao, J. Yang, Y. Chen, J. Yi, and H. Su, ``Quasi-direct drive actuation for a lightweight hip exoskeleton with high backdrivability and high bandwidth,'' \emph{IEEE/ASME Trans. Mechatron.}, vol. 25, pp. 1794--1802, 2020. \href{https://doi.org/10.1109/TMECH.2020.2995134}{doi: 10.1109/TMECH.2020.2995134}.

\bibitem{ref22} M. Grebenstein, M. Chalon, W. Friedl, S. Haddadin, T. Wimb\"ock, G. Hirzinger, and R. Siegwart, ``The hand of the DLR Hand Arm System: Designed for interaction,'' \emph{Int. J. Robot. Res.}, vol. 31, pp. 1531--1555, 2012. \href{https://doi.org/10.1177/0278364912459209}{doi: 10.1177/0278364912459209}.

\bibitem{ref23} K. Junge and J. Hughes, ``ADAPT-Teleop: Robotic hand with human matched embodiment enables dexterous teleoperated manipulation,'' \emph{npj Robot.}, vol. 3, art. no. 31, 2025. \href{https://doi.org/10.1038/s44182-025-00034-3}{doi: 10.1038/s44182-025-00034-3}.

\bibitem{ref24} C. Della Santina, C. Piazza, G. Grioli, M. G. Catalano, and A. Bicchi, ``Toward dexterous manipulation with augmented adaptive synergies: The Pisa/IIT SoftHand 2,'' \emph{IEEE Trans. Robot.}, vol. 34, pp. 1141--1156, 2018. \href{https://doi.org/10.1109/TRO.2018.2830407}{doi: 10.1109/TRO.2018.2830407}.

\bibitem{ref25} H. Li, C. J. Ford, C. Lu, Y. Lin, M. Bianchi, M. G. Catalano, E. Psomopoulou, and N. F. Lepora, ``Tactile SoftHand-A: 3D-printed, tactile, highly underactuated, anthropomorphic robot hand with an antagonistic tendon mechanism,'' \emph{Int. J. Robot. Res.}, vol. 45, pp. 1063--1090, 2026. \href{https://doi.org/10.1177/02783649251379516}{doi: 10.1177/02783649251379516}.

\bibitem{ref28} D. B. Camarillo, C. F. Milne, C. R. Carlson, M. R. Zinn, and J. K. Salisbury, ``Mechanics modeling of tendon-driven continuum manipulators,'' \emph{IEEE Trans. Robot.}, vol. 24, pp. 1262--1273, 2008. \href{https://doi.org/10.1109/TRO.2008.2002311}{doi: 10.1109/TRO.2008.2002311}.

\bibitem{ref29} S. Awtar, A. Kim, and A. Thombre, ``Tension management in cable transmission systems for remote manipulators,'' \emph{Precis. Eng.}, vol. 72, pp. 878--890, 2021. \href{https://doi.org/10.1016/j.precisioneng.2021.04.010}{doi: 10.1016/j.precisioneng.2021.04.010}.

\bibitem{ref30} H. Kawasaki, T. Komatsu, and K. Uchiyama, ``Dexterous anthropomorphic robot hand with distributed tactile sensor: Gifu Hand II,'' \emph{IEEE/ASME Trans. Mechatron.}, vol. 7, pp. 296--303, 2002. \href{https://doi.org/10.1109/TMECH.2002.802720}{doi: 10.1109/TMECH.2002.802720}.

\bibitem{ref32} A. Handa, A. Allshire, V. Makoviychuk, A. Petrenko, R. Singh, J. Liu, D. Makoviichuk, K. Van Wyk, A. Zhurkevich, B. Sundaralingam, Y. Narang, J.-F. Lafleche, D. Fox, and G. State, ``DeXtreme: Transfer of agile in-hand manipulation from simulation to reality,'' in \emph{Proc. IEEE Int. Conf. Robot. Autom. (ICRA)}, pp. 5977--5984, 2023. \url{https://doi.org/10.1109/ICRA48891.2023.10160216}.

\bibitem{ref35} Q. Li, H. Zhou, D. Mei, and Y. Wang, ``Highly efficient evaporative cooling of humanoid robot joint motors based on phase change hydrogels,'' \emph{Appl. Therm. Eng.}, vol. 298, art. no. 131111, 2026. \href{https://doi.org/10.1016/j.applthermaleng.2026.131111}{doi: 10.1016/j.applthermaleng.2026.131111}.

\bibitem{ref36} F. Yang, Y. Sun, J. Yang, S. Zhang, H. Liu, and Y. Tang, ``Effect of forced convection cooling on overload performance of permanent magnet synchronous motor for industrial robot,'' \emph{Int. Commun. Heat Mass Transf.}, vol. 161, art. no. 108527, 2025. \href{https://doi.org/10.1016/j.icheatmasstransfer.2024.108527}{doi: 10.1016/j.icheatmasstransfer.2024.108527}.

\bibitem{ref42} L. Birglen and C. M. Gosselin, ``Kinetostatic analysis of underactuated fingers,'' \emph{IEEE Trans. Robot. Autom.}, vol. 20, pp. 211--221, 2004. \href{https://doi.org/10.1109/TRA.2004.824641}{doi: 10.1109/TRA.2004.824641}.

\bibitem{ref43} T. Lalibert{\'e}, L. Birglen, and C. M. Gosselin, ``Underactuation in robotic grasping hands,'' \emph{Mach. Intell. Robot. Control}, vol. 4, pp. 77--87, 2002.

\bibitem{ref45} G. Rothenhofer, C. Walsh, and A. Slocum, ``Transmission ratio based analysis and robust design of mechanisms,'' \emph{Precis. Eng.}, vol. 34, pp. 790--797, 2010. \href{https://doi.org/10.1016/j.precisioneng.2010.03.010}{doi: 10.1016/j.precisioneng.2010.03.010}.

\bibitem{ref46} H. Yang, G. Wei, L. Ren, Z. Qian, K. Wang, H. Xiu, and W. Liang, ``An affordable linkage-and-tendon hybrid-driven anthropomorphic robotic hand---MCR-Hand II,'' \emph{J. Mech. Robot.}, vol. 13, art. no. 024502, 2021. \href{https://doi.org/10.1115/1.4049744}{doi: 10.1115/1.4049744}.

\bibitem{ref48} J. Zhou, J. Huang, Q. Dou, P. Abbeel, and Y. Liu, ``A dexterous and compliant (DexCo) hand based on soft hydraulic actuation for human-inspired fine in-hand manipulation,'' \emph{IEEE Trans. Robot.}, vol. 41, pp. 666--686, 2025. \href{https://doi.org/10.1109/TRO.2024.3508932}{doi: 10.1109/TRO.2024.3508932}.

\bibitem{ref50} R. Deimel and O. Brock, ``A novel type of compliant and underactuated robotic hand for dexterous grasping,'' \emph{Int. J. Robot. Res.}, vol. 35, pp. 161--185, 2016. \href{https://doi.org/10.1177/0278364915592961}{doi: 10.1177/0278364915592961}.

\bibitem{ref52} L. Heng, Y. Tang, J. Xu, H. Bao, D. Huang, and Y. Wang, ``HumDex: Humanoid dexterous manipulation made easy,'' arXiv preprint arXiv:2603.12260, 2026. \url{https://arxiv.org/abs/2603.12260}.

\bibitem{ref53} B. Zitkovich, T. Yu, S. Xu, P. Xu, T. Xiao, F. Xia, J. Wu, P. Wohlhart, S. Welker, A. Wahid, Q. Vuong, V. Vanhoucke, H. Tran, R. Soricut, A. Singh, J. Singh, P. Sermanet, P. R. Sanketi, G. Salazar, M. S. Ryoo, K. Reymann, K. Rao, K. Pertsch, I. Mordatch, H. Michalewski, Y. Lu, S. Levine, L. Lee, T.-W. E. Lee, I. Leal, Y. Kuang, D. Kalashnikov, R. Julian, N. J. Joshi, A. Irpan, B. Ichter, J. Hsu, A. Herzog, K. Hausman, K. Gopalakrishnan, C. Fu, P. Florence, C. Finn, K. A. Dubey, D. Driess, T. Ding, K. M. Choromanski, X. Chen, Y. Chebotar, J. Carbajal, N. Brown, A. Brohan, M. Gonzalez Arenas, and K. Han, ``RT-2: Vision-language-action models transfer web knowledge to robotic control,'' in \emph{Proc. Mach. Learn. Res.}, vol. 229, pp. 2165--2183, 2023. \url{https://arxiv.org/abs/2307.15818}.

\bibitem{ref54} M. J. Kim, K. Pertsch, S. Karamcheti, T. Xiao, A. Balakrishna, S. Nair, R. Rafailov, E. P. Foster, P. R. Sanketi, Q. Vuong, T. Kollar, B. Burchfiel, R. Tedrake, D. Sadigh, S. Levine, P. Liang, and C. Finn, ``OpenVLA: An open-source vision-language-action model,'' in \emph{Proc. Mach. Learn. Res.}, vol. 270, pp. 2679--2713, 2025. \url{https://proceedings.mlr.press/v270/kim25c.html}.

\bibitem{ref55} K. Black, N. Brown, D. Driess, A. Esmail, M. Equi, C. Finn, N. Fusai, L. Groom, K. Hausman, B. Ichter, S. Jakubczak, T. Jones, L. Ke, S. Levine, A. Li-Bell, M. Mothukuri, S. Nair, K. Pertsch, L. X. Shi, J. Tanner, Q. Vuong, A. Walling, H. Wang, and U. Zhilinsky, ``$\pi_0$: A vision-language-action flow model for general robot control,'' in \emph{Proc. Robotics: Science and Systems (RSS)}, 2025. \url{https://doi.org/10.15607/RSS.2025.XXI.010}.

\bibitem{ref56} M. Kallmann, ``Analytical inverse kinematics with body posture control,'' \emph{Comput. Animat. Virtual Worlds}, vol. 19, pp. 79--91, 2008. \href{https://doi.org/10.1002/cav.176}{doi: 10.1002/cav.176}.

\bibitem{ref57} P. Dobrowolski, ``Swing-twist decomposition in Clifford algebra,'' arXiv preprint arXiv:1506.05481, 2015. \url{https://arxiv.org/abs/1506.05481}.

\bibitem{ref58} T. Sebastian, ``Temperature effects on torque production and efficiency of PM motors using NdFeB magnets,'' \emph{IEEE Trans. Ind. Appl.}, vol. 31, pp. 353--357, 1995. \href{https://doi.org/10.1109/28.370284}{doi: 10.1109/28.370284}.

\bibitem{ref59} S. Li, B. Sarlioglu, S. Jurkovic, and N. R. Patel, ``Analysis of temperature effects on performance of interior permanent magnet machines for high variable temperature applications,'' \emph{IEEE Trans. Ind. Appl.}, vol. 53, pp. 4923--4933, 2017. \href{https://doi.org/10.1109/TIA.2017.2700473}{doi: 10.1109/TIA.2017.2700473}.

\bibitem{ref60} S. Park, S. Lee, M. Choi, J. Lee, J. Kim, J. Kim, and H. Joo, ``Learning to transfer human hand skills for robot manipulations,'' arXiv preprint arXiv:2501.04169, 2025. \url{https://arxiv.org/abs/2501.04169}.

\bibitem{ref62} Q. Mao, Z. Liao, J. Yuan, and R. Zhu, ``Multimodal tactile sensing fused with vision for dexterous robotic housekeeping,'' \emph{Nat. Commun.}, vol. 15, art. no. 6871, 2024. \href{https://doi.org/10.1038/s41467-024-51261-5}{doi: 10.1038/s41467-024-51261-5}.

\bibitem{ref63} A. Yamaguchi and C. G. Atkeson, ``Tactile behaviors with the vision-based tactile sensor FingerVision,'' \emph{Int. J. Humanoid Robot.}, vol. 16, art. no. 1940002, 2019. \href{https://doi.org/10.1142/S0219843619400024}{doi: 10.1142/S0219843619400024}.

\bibitem{ref64} S. Park and D. Hwang, ``Softness-adaptive pinch-grasp strategy using fingertip tactile information of robot hand,'' \emph{IEEE Robot. Autom. Lett.}, vol. 6, pp. 6370--6377, 2021. \href{https://doi.org/10.1109/LRA.2021.3092770}{doi: 10.1109/LRA.2021.3092770}.

\bibitem{ref65} A. H. A. Stienen, E. E. G. Hekman, F. C. T. van der Helm, G. B. Prange, M. J. A. Jannink, A. M. M. Aalsma, and H. van der Kooij, ``Freebal: Dedicated gravity compensation for the upper extremities,'' in \emph{Proc. IEEE Int. Conf. Rehabil. Robot. (ICORR)}, pp. 804--808, 2007. \url{https://doi.org/10.1109/ICORR.2007.4428517}.

\bibitem{ref66} S. U. Chon, J. Seo, J. Kim, S. Han, S. Park, J. T. Kim, J. Kim, and J. Cho, ``Design of an intuitive master for improving teleoperation task performance using the functional separation of actuators: Movement and gravity compensation,'' \emph{Actuators}, vol. 11, art. no. 204, 2022. \href{https://doi.org/10.3390/act11070204}{doi: 10.3390/act11070204}.

\bibitem{ref67} J. Atkins, D. Chang, and H. Lee, ``Design of a wearable shoulder exoskeleton robot with dual-purpose gravity compensation and a compliant misalignment compensation mechanism,'' \emph{Wearable Technol.}, vol. 5, art. no. e4, 2024. \href{https://doi.org/10.1017/wtc.2024.1}{doi: 10.1017/wtc.2024.1}.

\end{thebibliography}
\end{document}